\makeatletter\def\graphicscache@inhibit{true}\makeatother

\documentclass[letterpaper, 10 pt, conference]{ieeeconf}  %

\IEEEoverridecommandlockouts                              %

\usepackage{amsmath} %
\usepackage{amssymb}  %
\usepackage{nccmath}
\usepackage{graphicx}
\usepackage{booktabs}
\usepackage{threeparttable}
\usepackage{spreadtab}
\usepackage[nomargin,inline,draft]{fixme}

\usepackage{tikz}
\usepackage{pgfplots}
\usetikzlibrary{fit,shapes.callouts,shapes.geometric,backgrounds,positioning,arrows.meta}
\graphicspath{{./figures/}}

\usepackage{tensor}

\usepackage[hidelinks]{hyperref}
\usepackage[capitalize]{cleveref}
\crefname{section}{section}{sections}

\usepackage[style=ieee,hyperref,natbib=true,backend=bibtex,firstinits,doi=false,%
mincitenames=1,maxcitenames=2,maxbibnames=99,sorting=none,terseinits=false,hyperref=true]{biblatex}
\bibliography{references.bib}
\renewbibmacro*{bbx:savehash}{}%
\AtEveryBibitem{\clearfield{pages}\clearlist{publisher}\clearlist{organization}} %
\defbibheading{bibliography}[\bibname]{\section*{References}}

\usepackage{lipsum}
\usepackage[export]{adjustbox}

\newcommand{\E}[2]{\ensuremath{\tensor[^{#2}]{E}{_{#1}}}}

\usepackage{tikz}
\usetikzlibrary{arrows}
\usetikzlibrary{positioning,calc}
\usetikzlibrary{decorations.pathreplacing}
\usetikzlibrary{decorations.markings}
\usetikzlibrary{decorations.pathmorphing}
\usetikzlibrary{calligraphy}
\usetikzlibrary{fit}
\usetikzlibrary{shapes.arrows}
\usetikzlibrary{shapes.geometric}
\usetikzlibrary{matrix}
\usetikzlibrary{spy}

\pgfdeclarelayer{background}
\pgfdeclarelayer{foreground}
\pgfsetlayers{background,main,foreground}

\usepackage{pgfplots}
\usepackage{scalefnt}
\makeatletter

\tikzdeclarecoordinatesystem{rel}{%
	\tikzset{cs/.cd,x=0pt,y=0pt,#1}%
	\pgfpointlineattime{(\tikz@cs@x / 100)}%
	{\pgfpointanchor{\tikz@pp@name{\tikz@cs@node}}{south west}}%
	{\pgfpointanchor{\tikz@pp@name{\tikz@cs@node}}{south east}}%
	\edef\tikz@cs@x{\the\pgf@x}%
	\pgfpointlineattime{(\tikz@cs@y / 100)}%
	{\pgfpointanchor{\tikz@pp@name{\tikz@cs@node}}{south west}}%
	{\pgfpointanchor{\tikz@pp@name{\tikz@cs@node}}{north west}}%
	\pgfpoint{\tikz@cs@x}{\pgf@y}%
}

\makeatother

\IfFileExists{graphicscache.sty}{\usepackage[dpi=1200]{graphicscache}}

\title{\LARGE \bf
	Attention-Based VR Facial Animation with Visual Mouth Camera Guidance for Immersive Telepresence Avatars
}

\author{Andre Rochow, Max Schwarz, and Sven Behnke%
	\thanks{All authors are with the Autonomous Intelligent Systems group of University of Bonn, Germany; {\tt rochow@ais.uni-bonn.de}}%
}

\begin{document}

	\maketitle
	\thispagestyle{empty}
	\pagestyle{empty}

	\begin{abstract}
        Facial animation in virtual reality environments is essential for applications that necessitate clear visibility of the user's face and the ability to convey emotional signals.
		In our scenario, we animate the face of an operator who controls a robotic Avatar system. The use of facial animation is particularly valuable when the perception of interacting with a specific individual, rather than just a robot, is intended.
		Purely keypoint-driven animation approaches struggle with the complexity of facial movements. We present a hybrid method
		that uses both keypoints and direct visual guidance from a mouth camera.
		Our method generalizes to unseen operators and requires only a quick enrolment step with capture of two short videos.
        Multiple source images are selected with the intention to cover different facial expressions.
        Given a mouth camera frame from the HMD, we dynamically construct the target keypoints and apply an attention mechanism to determine the importance of each source image.
		To resolve keypoint ambiguities and animate a broader range of mouth expressions, we propose to inject visual mouth camera information into the latent space.
		We enable training on large-scale speaking head datasets by simulating the mouth camera input with its perspective differences
		and facial deformations.
		Our method outperforms a baseline in quality, capability, and temporal consistency.
		In addition, we highlight how the facial animation contributed to our victory at the ANA Avatar XPRIZE Finals.
	\end{abstract}

 	\section{Introduction}

    Facial animation is an important task in visual computing. A popular setting is face reenactment, where a source image and a driving image, which may be of different persons, are provided.
    The resulting image should have the appearance of the source image person, but the pose and facial expression of the driving image person.
    Generally, face reenactment methods are trained on speaking head datasets, such as Vox-Celeb~\cite{vox}. At inference time, the objective is to utilize arbitrary driving videos to animate the source-image person.

    A special case in virtual reality is VR facial animation, where the user wears a head-mounted display (HMD) and is, thus, not fully visible. Driving information has to be captured by sensors mounted on the headset and is typically incomplete. 
    Limited and occluded information together with large perspective offsets makes VR facial animation exceptionally challenging.
	Furthermore, many HMDs cause deformations in even the visible areas which particularly limits mouth movements.
    The alignment problem between mouth camera images and images without the presence of an HMD is one of the biggest challenges for generating training samples.

\begin{figure}[t]
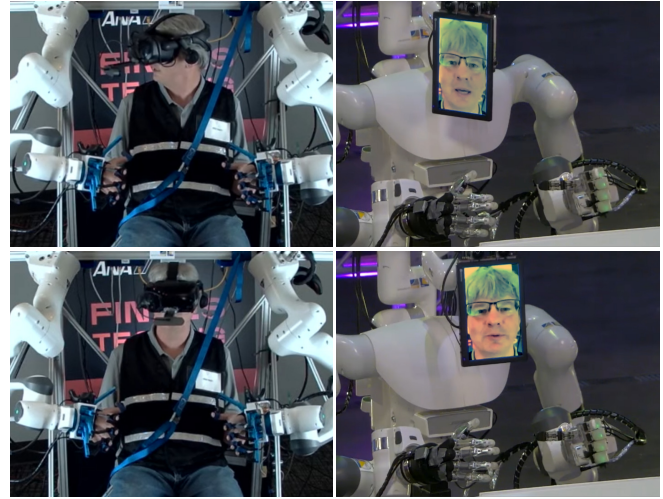

	\centering                                      %
	\includegraphics[width=0.497\linewidth,clip,trim=1cm 1cm 1cm 1cm]{Avatar_finals_images/4_op.png}\hfill
	\includegraphics[width=0.497\linewidth,clip,trim=0cm 0cm 3cm 3.15cm]{Avatar_finals_images/4_av_edited.png} \\
	\vspace{0.2em}
	\includegraphics[width=0.497\linewidth,clip,trim=1cm 1cm 1cm 1cm]{Avatar_finals_images/10_op.png}\hfill
	\includegraphics[width=0.497\linewidth,clip,trim=0cm 0cm 3cm 3.15cm]{Avatar_finals_images/10_av_edited.png} \\
	\vspace{-1ex}
	\caption[Op]{Facial animation of an operator interacting with a recipient through the NimbRo Avatar system at the ANA Avatar XPRIZE Finals.
		Stills from our winning run\footnotemark. Contrast was enhanced for easier viewing.}
	\label{fig:teaser}\vspace{-3ex}
\end{figure}

    Our system was developed for the ANA Avatar XPRIZE Challenge\footnote{\url{https://www.xprize.org/prizes/avatar}},
	where a previously unknown operator had to perform various tasks through our Avatar robot system (see \cref{fig:teaser}).
	Our robotic system~\citep{schwarz2021nimbro,nimbro2023_avatar} consists of an operator station with a VR headset and arm exoskeletons, as well as
	an avatar robot.
	We use a modified \textit{Valve Index} HMD equipped with two infrared eye cameras and a mouth camera~\citep{baseline}.
	As visualized in \cref{fig:teaser}, the operator's face is animated on a display that mirrors the operator head movement using an 6~DoF robotic arm.
	At the competition participants were judged not only on task performance, but also on immersion and the communication experience of
	a remote recipient. In particular, points were awarded when the operator was able to convey emotional cues to the recipient.
	Facial animation was thus a cornerstone of our strong performance at the
	challenge finals in November 2022, where our team NimbRo won the first prize.

\footnotetext{\url{https://www.youtube.com/watch?v=OD2UbZNw9sQ}}

    In our previous work~\cite{baseline}, we formulated VR facial animation as a keypoint-driven face reenactment problem.
    This allowed us to train on large speaking head datasets and leverage knowledge obtained from many different appearances to animate unseen persons. %
    In particular, we guided animation of the mouth area by dynamically retrieving a source image based on its keypoint similarity to the mouth camera image.
    Unfortunately, temporal inconsistencies occurred whenever changing the expression frame and the animation accuracy strongly depended on the image retrieval quality.
    Furthermore, keypoint ambiguities limit the range of possible expressions.

    In this work, we propose an extension of \cite{baseline} to address these limitations while preserving the ability to generalize to unseen persons.
    We propose to utilize multiple source images with an attention mechanism driven by the mouth camera, enabling our method to dynamically weight relevant features.
    Using the mouth camera video stream as the driving input reduces temporal inconsistencies, since the attention values are estimated by a continuous function that adapts to changes in the input.

	We enhance the range of possible facial expressions and solve keypoint ambiguities by introducing a mouth camera guidance that directly utilizes visual mouth camera features. As discussed, the alignment problem makes it very challenging to generate suitable training data.
	We adress this issue by proposing an efficient way to keep training on large speaking head datasets for generalizability during inference and additionally annotate a few image pairs with similar mouth expressions in the mouth and face camera that we merge into the training process.
	As we demonstrate in a detailed evaluation and the supplementary video\footnote{\scalefont{0.953}\url{https://www.ais.uni-bonn.de/videos/IROS_2023_Rochow}\label{supplementary_video}}, our VR facial animation pipeline generates more accurate and more temporally consistent results than the baseline, with more movement in areas that are not associated with keypoints, such as the cheeks.

	In addition to a real-time capable VR facial animation pipeline, our contributions include: (i) a source image attention mechanism that significantly improves temporal consistency and facial animation accuracy, (ii) an efficient way to leverage visual Mouth Camera information to resolve keypoint ambiguities and model a broader range of facial expressions, and (iii) emulation of mouth camera data, which allows
	training on available large-scale datasets.

	\section{Related Work}

	\subsection{Face Reenactment}
	A task related to VR facial animation is face reenactment. Here, a driving frame which encodes the head pose and expressional information is to be visualized with the appearance given by a source image person. Often keypoints are used to represent the motion~\cite{monkey_net,fom,zhao2021sparse}. A motion network predicts a deformation grid to deform source images into a defined target motion. \citet{fom} propose to use image feature-based local affine transformations in the motion network that allow to model a larger family of transformations. \citet{face_nerf} propose to condition a dynamic NeRF with motion information extracted from driving images.
	
	\subsection{VR Facial Animation}
	In VR facial animation, the motion is encoded in eye camera images and a mouth camera image \cite{lombardi2018deep,vr_facial, baseline} or even in audio  recordings~\cite{faceaudio}.
	\citet{lombardi2018deep} render a virtual avatar by utilizing a variational autoencoder (VAE) that can be conditioned with motion parameters obtained from the HMD. They train a second VAE on real and synthetic mouth camera images and map similar expressions of both domains to similar latent codes by manually controlling the latent variable that determines the domain.
	However, they do not handle facial deformations caused by the HMD explicitly.
	\citet{vr_facial} generate synthetic ground truth data with an expression-preserving style transfer network, which maps between the mouth camera domain and the avatar domain.
	\citet{faceaudio} bypass the alignment problem by omitting mouth camera images completely and using audio instead. They generate impressive results; however, the reduced amount of information significantly limits the expressivity. Especially when the user is silent, the animation task is ill-posed.
	Unfortunately, all these methods need a significant amount of data capture and operator-specific training, which makes them unsuitable for use-cases that require instant application, such as the ANA Avatar XPRIZE Competition.
	
	\begin{figure}
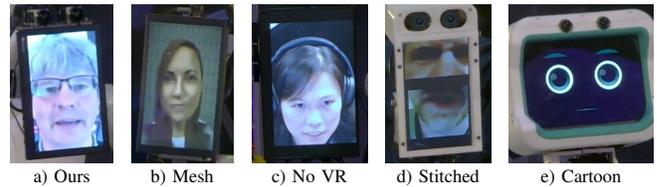
\centering
 \begin{tikzpicture}[
    every label/.style={font=\scriptsize},
    every node/.style={inner sep=0pt},
    label distance=.5ex
  ]
  \node[matrix, column sep=7pt] {
    \node[label={south:a) Ours}] {\includegraphics[height=2.1cm]{Avatar_finals_images/comp/nimbro.png}}; &
    \node[label={south:b) Mesh}] {\includegraphics[height=2.1cm]{Avatar_finals_images/comp/pollen.png}}; &
    \node[label={south:c) No VR}] {\includegraphics[height=2.1cm]{Avatar_finals_images/comp/northeastern.png}}; &
    \node[label={south:d) Stitched}] {\includegraphics[height=2.1cm]{Avatar_finals_images/comp/avatrina.png}}; &
    \node[label={south:e) Cartoon}] {\includegraphics[height=2.1cm,clip,trim=40px 0px 40px 0px]{Avatar_finals_images/comp/unist.png}}; \\
  };
 \end{tikzpicture} 
 \vspace{-4ex}
 \caption{Types of facial animation at ANA Avatar XPRIZE finals. Examples from teams: a) NimbRo (first place),
 b)~Pollen Robotics~(second), c)~Northeastern~\cite{luo2022towards}~(third), d)~AVATRINA~\citep{marques2022commodity}~(fourth), e)~UNIST (sixth).}
 \label{fig:types} \vspace{-2ex}
\end{figure}
	
	From ANA Avatar XPRIZE finals video footage we recognize five categories of face animation techniques used by participants (see \cref{fig:types}).
	Out of the 12 teams selected for the two competition days, three teams had no VR headset. In this case, video streaming suffices for face display, but operator immersion is limited. Very similarly, one team displayed mouth camera footage directly, stitched together with previously captured footage of the operator's eyes.
The rest of the teams used expression information from mouth trackers and/or audio to animate either 2D emoji drawings (three teams) or rendered 3D meshes, adapted to roughly match the operator's attributes like hair color and gender (four teams).
Our team was the only one to produce a photorealistic animated face image.

	\begin{figure*}[htb!]
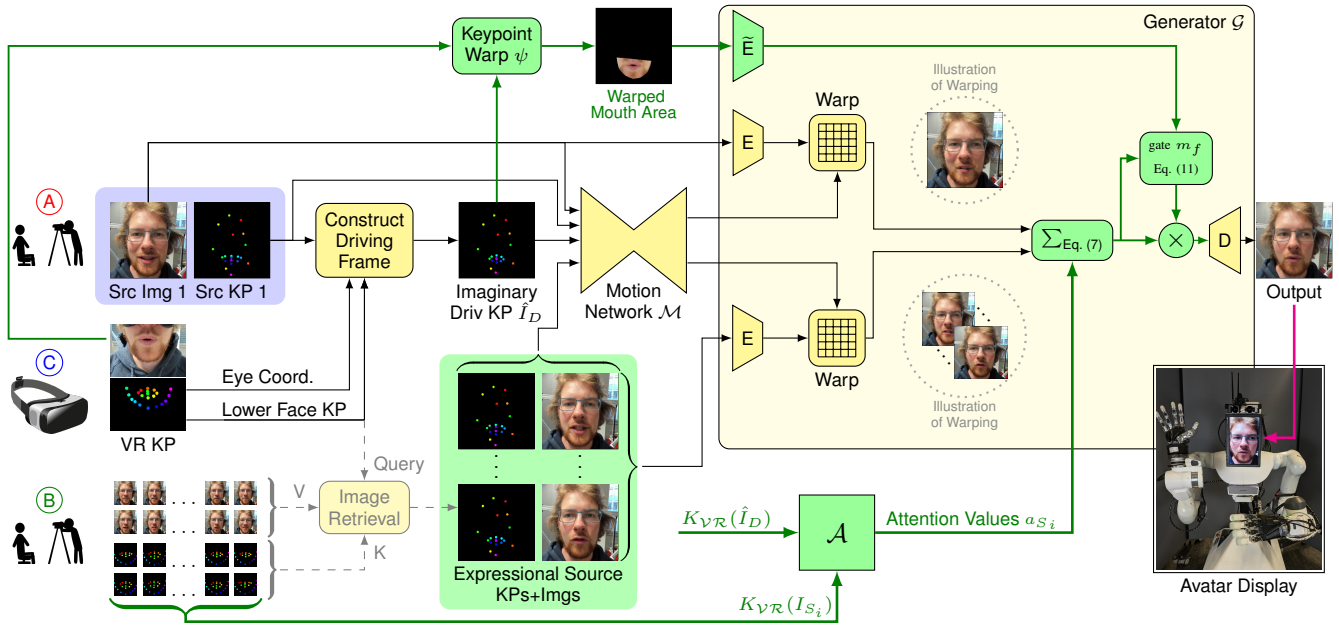

	\centering
	\begin{tikzpicture}[
	font=\sffamily\scriptsize,
	inp/.style={draw,minimum width=1cm,minimum height=1cm,inner sep=0},
	inp_c/.style={draw, dotted,circle,thick,gray!70,minimum width=1cm,minimum height=1cm,inner sep=0},
	inp_nb/.style={draw=none,minimum width=1cm,minimum height=1cm,inner sep=0},
	inp_half/.style={draw=none,minimum width=1cm,minimum height=0.5cm,inner sep=0},
	inp_sm/.style={draw=none,minimum width=0.7cm,minimum height=0.7cm,inner sep=0},
	m/.style={draw,rounded corners,fill=yellow!50,align=center},
	mg/.style={draw,rounded corners,fill=green!40,align=center},
	node distance=0.6,
	every label/.style={font=\sffamily\scriptsize,align=center,inner sep=2pt,text depth=1pt},
	label distance=0pt,
	]
	
	\node[inp,label=south:Src Img 1] (source_img) {\includegraphics[width=1cm]{images_2023/fig_infer/chris/193/source_0.png}};

	\node[inp,right=0.1 of source_img,label=south:Src KP 1] (source_kp) {\includegraphics[width=1cm]{images_2023/fig_infer/chris/193/kp_source_s4_0.png}};
	
	\node[inner sep=0,left=.3cm of source_img,label={[red,draw,circle,text depth=0,inner sep=1pt]north:A}] (portrait) {
		\includegraphics[width=1cm,clip,trim=20 0 20 0]{images/portrait.png}
	};
	
	\node[m,right=of source_kp] (constr) {Construct\\Driving\\Frame};
	
	\node[inp,right=of constr,label={south:Imaginary\\Driv KP $\hat{I}_D$}] (driving_kp) {\includegraphics[width=1cm]{images_2023/fig_infer/chris/193/kp_driving_s4.png}};
	
	\node[m,above=1.7cm of driving_kp,fill=green!40] (delauny_transform) {Keypoint\\Warp $\mathbf{\psi}$};
	
	\node[inner sep=0,right=of driving_kp,label={[yshift=0.6em]south:Motion\\Network $\mathcal{M}$}] (motion) {\tikz[scale=0.7]{
			\draw[fill=yellow!50] (-1,-1) -- (-1,1) -- (0,0.2) -- (1,1) -- (1,-1) -- (0,-0.2) -- cycle;}
	};

	\node[draw,fill=yellow!50,trapezium,shape border rotate=270] at ($(motion.east)+(0.8,-1.3)$) (expr_enc) {E};
	\node[m,anchor=west,minimum height=0.5cm,label=south:Warp] at ($(expr_enc.east)+(0.6,0)$) (warp_expr) {\tikz{
			\draw[step=0.1,black,thin] (0.0,0.0) grid (0.5,0.5);
	}};

	\node[inp_sm] at ($(warp_expr.east)+(+1.1,0.25)$) (deformed_ex1) {\includegraphics[width=0.7cm]{images_2023/fig_infer/chris/193/deformed_2.png}};
	\node[inp_sm] at ($(deformed_ex1.north)+(0.43,-0.8)$) (deformed_ex2) {\includegraphics[width=0.7cm]{images_2023/fig_infer/chris/193/deformed_4.png}};
	\node[right=0.2,inner sep=1pt,rotate=-50] at ($(deformed_ex1.north)+(0.15,-0.15)$) (dots1) {\dots};
	\node[right=0.2,inner sep=1pt,rotate=-50] at ($(deformed_ex1.south)+(-0.38,0)$) (dots2) {\dots};
	\node[draw,dotted,circle,fit=(deformed_ex1)(deformed_ex2),inner sep=0, gray!70, thick,label={[name=label_deformed_expr,yshift=0.1em]south:\textcolor{gray!90}{\scalefont{0.75}Illustration}\\[-0.07cm] \textcolor{gray!90}{\scalefont{0.75}of Warping}}] (deformed_expr) {};

	\node[draw,fill=yellow!50,trapezium,shape border rotate=270,anchor=west] at ($(motion.east)+(0.6,1.3)$) (source_enc) {E};
	
	\node[m,anchor=west,minimum height=0.5cm,label=north:Warp] at ($(source_enc.east)+(0.6,0)$) (warp_source) {\tikz{
			\draw[step=0.1,black,thin] (0.0,0.0) grid (0.5,0.5);
	}};
	\node[inp_c,label={[yshift=-0.1em]north:\textcolor{gray!90}{\scalefont{0.75}Illustration}\\[-0.07cm] \textcolor{gray!90}{\scalefont{0.75}of Warping}}]  at ($(warp_source.east)+(+1.31,-0.1)$) (deformed_source) {\includegraphics[width=1cm]{images_2023/fig_infer/chris/193/deformed_0.png}};

	\node[m,anchor=west,fill=green!40] at ($(warp_expr.east)!0.5!(warp_source.east)+(2.2,0)$) (sum) {$\sum_{\text{\cref{eq:attention}}}$};
	\node[m, circle, fill=green!40,font=\normalsize,inner sep=1pt,right=0.6cm of sum] (odot) {$\times$};
	\node[m,font = \normalsize,above=0.5cm of odot,fill=green!40] (gating) {\tiny{gate $m_f$} \\[-0.1cm] \tiny \cref{eq:gating}};
	\node[inp,label=south:{{\scalefont{0.9} \textcolor{green!50!black}{Warped}}\\[-0.07cm] \textcolor{green!50!black}{{\scalefont{0.9}Mouth Area}}}] at (delauny_transform -| motion) (delauny_mouth) {\includegraphics[width=1cm]{images_2023/fig_infer/chris/193/delauny.png}};
	\node[draw,fill=green!40,trapezium,shape border rotate=270] at (source_enc |- delauny_mouth) (mouth_enc) {$\widetilde{\text{E}}$};
	\node[draw,fill=yellow!50,trapezium,shape border rotate=90,right=0.2 of odot] (dec) {D};
	\node[inp,right=0.2 of dec,label=south:Output] (out) {\includegraphics[width=1cm]{images_2023/fig_infer/chris/193/out.png}};
	
	\begin{pgfonlayer}{background}
	\node[fit=(sum)(expr_enc)(source_enc)(dec)(mouth_enc)(deformed_expr)(label_deformed_expr),draw=black,rounded corners,inner sep=5pt,fill= yellow!12] (g) {};
	\end{pgfonlayer}
	\node[anchor=north east,align=right] at (g.north east) {Generator $\mathcal{G}$};

	\node[inp,fill=green!40,font=\normalsize,shape border rotate=90,below=1.7 of warp_expr] (attention) {$\mathcal{A}$};

	\node[inp_half,below=0.6cm of source_img,label] (vr_img) {\includegraphics[width=1cm, clip,trim=1.25cm 0cm 1.25cm 0cm]{images_2023/fig_infer/chris/193/vr.png}};
	\node[inp_half,below=1.33cm of source_img,label=south:VR KP] (vr_kp) {\includegraphics[width=1cm, clip,trim=1.25cm 0cm 1.25cm 0cm]{images_2023/fig_infer/chris/193/kp_vr_s4.png}};
	\coordinate (projin) at ($(vr_kp.east)+(0,-0.2)$);
	\coordinate[m,right=0.5 of projin] (project); %
	
	\node[inner sep=0,left=.3cm of vr_kp,label={[blue,draw,circle,text depth=0,inner sep=1pt]north:C}] (vr_headset) {
		\includegraphics[width=1cm]{images/vr_headset.png}
	};

	\node[m,below=2.7cm of constr,text=gray,draw=gray,fill=yellow!30] (retrieval) {Image\\Retrieval};
	
	\node[anchor=west,matrix,matrix of nodes,inner sep=0,nodes={inner sep=1pt,minimum width=0.4cm,text depth=-5pt},row sep=0.5pt, column sep=-.5pt]   at ($(vr_kp.west|-retrieval.west)+(0,-0.83)$) (keys) {
		{\includegraphics[width=0.3cm]{images_2023/fig_infer/chris/kp_vr_s4__0.png}} & {\includegraphics[width=0.3cm]{images_2023/fig_infer/chris/kp_vr_s4__24.png}}  &
		{$\dots$}& {\includegraphics[width=0.3cm]{images_2023/fig_infer/chris/kp_vr_s4__48.png}} &
		{\includegraphics[width=0.3cm]{images_2023/fig_infer/chris/kp_vr_s4__72.png}} \\
		{\includegraphics[width=0.3cm]{images_2023/fig_infer/chris/kp_vr_s4__96.png}} & {\includegraphics[width=0.3cm]{images_2023/fig_infer/chris/kp_vr_s4__120.png}}  &
		{$\dots$}& {\includegraphics[width=0.3cm]{images_2023/fig_infer/chris/kp_vr_s4__144.png}} & {\includegraphics[width=0.3cm]{images_2023/fig_infer/chris/kp_vr_s4__298.png}} \\
	};
	\draw[pen colour={gray}, decorate, decoration = {calligraphic brace}, ultra thick] ($(keys.north east)+(0.1,0)$) -- ($(keys.south east)+(0.1,0)$);
	
	\node[anchor=west,matrix,matrix of nodes,inner sep=0,nodes={inner sep=1pt,minimum width=0.4cm,text depth=-5pt},row sep=0.5pt, column sep=-.5pt] at (vr_kp.west|-retrieval.west) (values) {
		{\includegraphics[width=0.3cm]{images_2023/fig_infer/chris/face_0.png}} & {\includegraphics[width=0.3cm]{images_2023/fig_infer/chris/face_24.png}} &
		{$\dots$}& {\includegraphics[width=0.3cm]{images_2023/fig_infer/chris/face_48.png}} & {\includegraphics[width=0.3cm]{images_2023/fig_infer/chris/face_72.png}} \\
		{\includegraphics[width=0.3cm]{images_2023/fig_infer/chris/face_96.png}} & {\includegraphics[width=0.3cm]{images_2023/fig_infer/chris/face_120.png}}  &
		{$\dots$}& {\includegraphics[width=0.3cm]{images_2023/fig_infer/chris/face_144.png}} & {\includegraphics[width=0.3cm]{images_2023/fig_infer/chris/face_298.png}} \\
	};
	\draw[pen colour={gray},decorate, decoration = {calligraphic brace},ultra thick] ($(values.north east)+(0.1,0)$) -- ($(values.south east)+(0.1,0)$);
	
	\coordinate (storage) at ($(keys.west)!0.5!(values.west)$);
	\node[inner sep=0,left=.3cm of storage,label={[green!50!black,draw,circle,text depth=0,inner sep=1pt]north:B}] (portrait2) {
		\includegraphics[width=1cm,clip,trim=20 0 20 0]{images/portrait.png}
	};
	
	\node[inp] at ($(retrieval.east)+(1.15,-0.2)$) (retrieved_kp) {\includegraphics[width=1cm]{images_2023/fig_infer/chris/193/kp_source_s4_4.png}};
	\node[inp,right=0.1 of retrieved_kp] (retrieved_img){\includegraphics[width=1cm]{images_2023/fig_infer/chris/193/source_4.png}};
	\node[every label,anchor=north] at ($(retrieved_kp.south)!0.5!(retrieved_img.south)$) {Expressional Source \\ KPs+Imgs};
	
	\node[inp] at ($(retrieved_img)+(0,1.48)$) (source_img2){\includegraphics[width=1cm]{images_2023/fig_infer/chris/193/source_2.png}};
	\node[] at ($(retrieved_img)+(0,0.85)$) (source_dots){$\vdots$};
	
	\node[inp] at ($(retrieved_kp)+(0,1.48)$) (source_kps2){\includegraphics[width=1cm]{images_2023/fig_infer/chris/193/kp_source_s4_2.png}};
	\node[] at ($(retrieved_kp)+(0,0.85)$) (source_dots2){$\vdots$};
	
	\draw[black,decorate,decoration={calligraphic brace,amplitude=6pt}, thick] ($(source_kps2.north west)+(-0.5pt,1pt)$)--($(source_img2.north east)+(0.5pt,1pt)$) node[midway, above](brace){};
	\node[draw,inner sep=1pt,anchor=north east,label=south:Avatar Display] (display) at ($(out.south east)+(-0.1,-1.15)$)
	{\includegraphics[width=2.2cm,clip,trim=0 240 50 160]{images_2023/avatar_display2.jpg}};
	
	\draw[black,decorate,decoration={calligraphic brace,amplitude=6pt}, thick] ($(source_img2.north east)+(1pt,0pt)$) -- ($(retrieved_img.south east)+(1pt,0pt)$) node[midway, above](brace2){};
	
	\draw[pen colour={black!50!green},decorate,decorate,decoration={calligraphic brace,amplitude=6pt},ultra thick] ($(keys.south east)+(0pt,-0.1pt)$) -- ($(keys.south west)+(0pt,-0.1pt)$) node[midway, above](brace_kp){};

	\begin{scope}[-latex]
	\draw (source_img.north) |- ($(source_enc.west)+(0,0)$);
	\draw (source_img.north) -- ($(source_img.north|-warp_source.west)+(0,0)$) -| ($(motion.west)+(-0.2,0.4)$) -- ($(motion.west)+(0,0.4)$);
	\draw (source_kp) -- (constr);
	\draw (source_kp.east) -- ++(0.3,0) -- ++(0,0.8) -| ($(driving_kp.east)!0.5!(motion.west)+(0,0.2)$) -- ($(motion.west)+(0,0.2)$);
	\draw (constr) -- (driving_kp);
	\draw (driving_kp) -- (motion);
	\draw[line width=0.25mm, black!50!green] (driving_kp) -- (delauny_transform);
	\draw[line width=0.25mm, black!50!green] (delauny_transform) -- (delauny_mouth);
	\draw[line width=0.25mm, black!50!green] (delauny_mouth) -- (mouth_enc);
	\draw[line width=0.25mm, black!50!green] (mouth_enc) -| (gating.north);
	\draw[line width=0.25mm, black!50!green] ($(vr_img.west)+(-0.05,0.15)$) --++(-1.3,0) |- (delauny_transform.west);
	\draw ($(motion.east)+(0,0.3)$) -- ++(0.2,0) -|  (warp_source.south);
	\draw (source_enc) -- (warp_source);
	\draw (warp_source.east) --++ (0.1,0)|- ($(sum.west)+(0,+0.15)$); 
	\draw[line width=0.25mm, black!50!green] ($(sum.east)+(0.1,0.0)$) |- (gating.west);
	
	\draw ($(motion.east)+(0,-0.3)$) -- ++(0.2,0) -|  (warp_expr.north);
	\draw (expr_enc) -- (warp_expr);

	\draw (warp_expr.east) --++ (0.1,0)|- ($(sum.west)+(0,-0.15)$); 
	\draw[line width=0.25mm, black!50!green] (sum) -- (odot);
	\draw[line width=0.25mm, black!50!green] (gating.south) -- (odot.north);
	\draw[line width=0.15mm, black!50!green] (odot) -- (dec);
	\draw (dec) -- (out);
	\draw[magenta, line width=0.35mm] (out.south) ++(0,-0.35) |- (rel cs:x=63,y=65,name=display);

	\draw[line width=0.35mm, black!50!green] (attention) -| (sum.south) node [black,pos=0.245,above,inner sep=1pt,black!50!green] {Attention Values $\medmath{{a_S}_i}$};

	\draw (vr_kp.east|-project) -| (constr);
	\draw ($(vr_kp.east)+(0,0.2)$) -| ($(constr.south)+(-0.2,0)$) node [pos=0.25,above,inner sep=1pt] {Eye Coord.};
	\draw (project) -| (constr) node [pos=0.215,above,inner sep=1pt]{Lower Face KP};
	\draw [dashed, gray](project -| constr) -- (retrieval) node [pos=0.75,right] {Query};
	\draw [dashed,gray] (values.east) ++(0.25,0) -- (retrieval) node [midway,above] {V};
	\draw [dashed,gray] (keys.east) ++(0.25,0) -| (retrieval) node [pos=0.75,right] {K};
	\draw [dashed,gray] (retrieval) -- ($(retrieved_kp.west)+(0,0.2)$);
	\draw (brace.north) -- ++(0,0.3) -| ($(driving_kp.east)!0.5!(motion.west)+(0,-0.3)$) -- ($(motion.west)+(0,-0.3)$);
	\draw ($(brace2.east)+(0.15,-0.125)$) -| ($(expr_enc.west)+(-0.4,0)$) -- ($(expr_enc.west)+(0,0)$);
	\draw [line width=0.35mm, black!50!green] ($(brace_kp.south)+(0,-0.2)$) -- ($(brace_kp.south)+(0,-0.3)$) -| (attention.south) node [black,left,inner sep=1pt,black!50!green,pos=0.65] {$K_{\mathcal{VR}}(I_{S_i})$};
	\draw [line width=0.35mm, black!50!green] ($(attention.west)+(-1.6,0)$) -- (attention) node [black,pos=0.4,above,inner sep=1pt,black!50!green] {$K_{\mathcal{VR}}(\hat{I}_{D})$};

	\end{scope}

	\begin{pgfonlayer}{background}
	\draw[fill=blue!18,rounded corners,draw=none]
	($(source_img.north west)+(-0.2,0.2)$) -|
	($(source_kp.south east)+(0.2,-0.35)$) -|
	($(source_img.south west)+(-0.2,-0.35)$)-- cycle;
	\end{pgfonlayer}
	
	\begin{pgfonlayer}{background}
	\draw[fill=green!30,rounded corners,draw=none]
	($(source_kps2.north west)+(-0.25,0.25)$) -|
	($(retrieved_img.south east)+(0.25,-0.65)$) -|
	($(retrieved_kp.south west)+(-0.25,-0.65)$)-- cycle;
	\end{pgfonlayer}
	\end{tikzpicture}
	
	\vspace{-1.5ex}
	\caption{Inference pipeline for VR Facial Animation. New components compared to our previous work~\cite{baseline} are highlighted in green. We select 4-5 still source images from a portrait video of the operator shot before the run as source images \textcolor{red}{(A)}. The remaining frames are optionally used as a key-value storage of retrievable expression keypoints and
		corresponding images \textcolor{green!50!black}{(B)}.
		The live keypoints measured inside and outside the VR headset \textcolor{blue}{(C)} are then projected to the first
		source image frame, where they are optionally used to retrieve the closest expression image with keypoints from the storage.
		The keypoints of all source images including the retrieved one and a constructed set of driving keypoints then
		enter the motion network $\mathcal{M}$, which estimates a deformation grid that is used to warp the source images features, extracted by the generator-encoder network, to match the driving keypoints. The illustration of warping in $\mathcal{G}$ shows the deformation grid applied to the image instead of encoded features.
		The deformed features are aggregated over the source images in the lower facial area using a trainable attention mechanism $\mathcal{A}$.
		The mouth camera image from the HMD is warped into the lower facial area of the constructed driving keypoints and then encoded by a separate encoder network $\widetilde{E}$. An estimated mask $m_f$ gates the aggregated deformed source features using the warped mouth camera features.
		The masked aggregated features are then decoded to produce the output.
	}
	\label{fig:infer}
	\vspace{-2.5ex}
\end{figure*}

	\section{Baseline} \label{sec:baseline}
	We select our avatar robot facial animation method developed in previous work~\citep{baseline} as a baseline, since it
	has the same input requirements and can thus be easily compared.

	We give a brief overview of our previous work here. It is composed of
    1) capturing and preprocessing, 2) image retrieval, 3) construction of the driving keypoints, 4) deforming, and 5) fusing and refining.

	\subsubsection{Capturing and Preprocessing} \label{preprocessing}
	We capture two videos of the operator, with and without the HMD, respectively. The mouth camera only captures the lower facial area and the second (source image) video captures the complete frontal facing head of the operator.
	From the source video, we select an arbitrary source image which subsequently defines the operator appearance.
	Keypoints are extracted from selected frames showing different facial expressions.
	We differentiate between lower facial area keypoints $\mathcal{K}_{VR}$, which are also visible in the mouth camera, and facial keypoints $\mathcal{K}_F$ that determine the head pose including a keypoint $kp_{eye}$ for the gaze direction.
	Given the set of mouth video keypoints $\medmath{\{\mathcal{K}_{VR}(I_{M_0}), \mathcal{K}_{VR}(I_{M_1}),\dots ,\mathcal{K}_{VR}(I_{M_k})\}}$ and source video keypoints $\medmath{\{\mathcal{K}_{VR}(I_{S_0}), \mathcal{K}_{VR}(I_{S_1}),\dots ,\mathcal{K}_{VR}(I_{S_l})\}}$, we define a keypoint mapping $\Pi_{S_i}(\mathcal{K}_{VR})$  that maps lower facial keypoints from the mouth camera into source image $I_{S_i}$.
	$\Pi(\cdot)$ corrects effects caused by the perspective change and deformations caused by the HMD's weight.

	\subsubsection{Image Retrieval} \label{base:retrieval}
	The image retrieval process searches the source video for a so-called \textit{expression image} that has a similar mouth expression as the live mouth camera image. Given the projection $\Pi_{S_i}(\mathcal{K_{VR}})$ of the mouth camera keypoints, we therefore retrieve the source video image $I_{S_i}$ with the best matching keypoints.
	The expression image is then utilized to guide the animation process in the lower facial area.
	
	\subsubsection{Construction of the Driving Keypoints}
	The keypoints $\mathcal{K}(\hat{I}_D)$ of an imaginary driving frame that fully specify the facial target expression and pose are constructed here.
	The 3D keypoints that determine the head pose are simply copied from the source image $\mathcal{K}_F(I_S) = \mathcal{K}(\hat{I}_D)$ since we move the face display on the robot and thus do not require head movement in the output animation.
	The gaze keypoints are estimated by transferring eye tracking results from the eye cameras into a normalized gaze coordinate system in the source image. For a detailed explanation we refer to \citet{baseline}.
    The lower facial keypoints $\mathcal{K}_{VR}(\hat{I}_D)$ are generated by projecting the current mouth camera keypoints into the fixed source image. We thus define the imaginary driving keypoints
    \begin{align}
	\mathcal{K}(\hat{I}_D) := \Pi_S (\,\mathcal{K}_{\mathcal{VR}}(I_M)\,) \oplus \rho(\,\mathcal{K}_{\mathcal{F}}(I_S), \hat{kp}_{eye}\,),
	\end{align}
	where $I_M$ is the mouth camera image, $\Pi_S(\cdot)$ maps each lower-face keypoint $kp_M^{(i)} \in \mathcal{K}_{\mathcal{VR}}(I_M)$ into the source image $I_S$, and
	$\rho(\cdot)$ replaces the eye keypoints detected in $I_S$ with the modified values $\hat{kp}_{eye}$ in order to include the operator's current gaze direction and eye openness.
	
	\subsubsection{Deforming}
	The motion network $\mathcal{M}$ generates deformation grids $\mathcal{M}_{S\leftarrow D}$ and $\mathcal{M}_{E\leftarrow D}$ that are used to sample a deformation of the source and expression image into the imaginary driving keypoints. The motion network cannot generate new content, but it generates a good initialization for the refinement network.
	
	\subsubsection{Fusing and Refining}
	The refinement network $\mathcal{G}$ combines the deformed source image and the lower facial area of the deformed expression image. It generates a realistic output image with the appearance of the source image and the facial expressions as specified by the constructed imaginary driving keypoints.

	\section{Method}
	Our proposed method is an extension of our previous work~\cite{baseline}. The basic modules and steps (see \cref{sec:baseline}) remain, with important functionalities added into the pipeline and refinement network.
	This extended refinement network is called generator $\mathcal{G}$ (see \cref{fig:infer,fig:train}).

	\subsection{Source Image Attention Mechanism}
	We address temporal inconsistencies, as occurring in our baseline method, by using more than two source images and introducing an attention mechanism that equips the network with the ability to decide on how much information it requires from each source view. 
	The attention mechanism works in several stages. We distinguish between two types of input images, the appearance (or first) source image $I_{S_1}$ and the expressional source images $I_{S_2},I_{S_3},\dots,I_{S_n}$.
	The first source image conserves all the appearance information of the operator, whereas the expressional source images are used to generate more accurate animations, by presenting the network different variations of the lower facial area of an operator.
	Especially the mouth area has a lot of variations due to occlusions, disocclusions and a many degrees of freedom when speaking.
	Given the selected source images $\medmath{I_{S_1}, I_{S_2}, \dots I_{S_n}}$ and the corresponding facial keypoints $\medmath{\mathcal{K}(I_{S_1}), \mathcal{K}(I_{S_2}), \dots, \mathcal{K}(I_{S_n})}$ we extract all keypoint sequences that correspond to the lower facial area $\medmath{\mathcal{K}_{VR}(I_{S_1}), \mathcal{K}_{VR}(I_{S_2}), \dots, \mathcal{K}_{VR}(I_{S_n})}$, which we call VR keypoints as they are also visible in the mouth camera of the HMD.
	For a sequence of VR keypoints $ kp_j \in \mathcal{K}_{VR}(I_{S_i})$ of the source image $I_{S_i}$, we generate a distance tensor $\mathcal{D}_{S_i}$ with
	\begin{align}
	\mathcal{D}_{S_i}^{k,l} = \frac{kp_k - kp_l}{\max \mathcal{D}_{S_i}} \in \mathbb{R}^2.
	\end{align}
	The distance tensor $\mathcal{D}_D$ is generated for the driving keypoints $\medmath{\mathcal{K}_{VR}(I_{D})}$ analogously.
    We then estimate similarity vectors of the source distance tensors
	\begin{align}
	\vec{x}_{S_i} &=  \vec{\mathcal{D}_{S_i}}W^S \in \mathbb{R}^{256}
	\end{align}
	and the driving distance tensor
	\begin{align}
	\vec{x}_D &=  \vec{\mathcal{D}_D}W^D  \in \mathbb{R}^{256},
	\end{align}
	where $W^S,W^D$$\in$ $\mathbb{R}^{d,256}$ are learned weight matrices and $\vec{\mathcal{D}}$ represents a flattened vector representation of a distance tensor.
    The similarity values are finally given by the scaled dot products
    \begin{align}
    x_{S_i} = \frac{\vec{x}_{S_i} \, \vec{x}_D^{\,\,\,\,T}}{\sqrt{256}} \in \mathbb{R},
    \end{align}
    which are fed into a softmax function to generate attention values $a_{S_i} \in \mathbb{R}$.
    These steps are summarized with $\mathcal{A}$ in \cref{fig:infer,fig:train}.
    
    Before we calculate the weighted sum we extract features $E_{S_i} = E(I_{S_1})$ of all source images $I_{S_i}$, using the generator encoder network $E$ (see \cref{fig:infer}), and align them in the driving keypoints. This is achieved by deforming the features into the driving keypoints using the deformation grid $\mathcal{M}_{S_i \leftarrow D}$ estimated by the motion network. The deformation generates a roughly aligned feature representation     
    \begin{align}
    \E{S_i}{D} &= \mathcal{M}_{S_i \leftarrow D}[\,E(I_{S_1})\,].
    \end{align}
    The aggregated deformed source image features
    \begin{align} \label{eq:attention}
    \E{S}{D} &= (1-B_{LF})\E{S_1}{D} + \sum_{i=1}^n a_{S_i}B_{LF} \E{S_i}{D},
	\end{align}
	are generated by a weighted sum in the lower facial area,
    where $B_{LF}$ is a binary mask that crops out the lower facial area of the deformed source images features and $a_{S_i}$ are the attention values.

	\subsection{Visual Mouth Camera Guidance} \label{guidance}
	We address keypoint ambiguities by leveraging visual information from the current mouth camera image to guide the animation process.
	It is challenging to directly process the mouth camera image due to perspective changes and deformations caused by the HMD.
	Our key idea for addressing this issue is to reuse the obtained lower facial (VR) keypoints from the mouth camera image $\mathcal{K}_{VR}(I_M)$ and its deformation-aware projection $\Pi_S(\mathcal{K}_{VR}(I_M))$ into the driving head pose. 
	We first estimate a Delaunay triangulation and then use barycentric coordinates to sample the mouth camera image in the target keypoints. We define the mouth area keypoint warping
	\begin{align}
	 \psi(\,I_1,\medmath{\mathcal{K}_{VR}(I_1)},\medmath{\mathcal{K}_{VR}(I_2)}\,) 
	\end{align} to be a function that samples the image $I_1$ with keypoints $\mathcal{K}_{VR}(I_1)$ in the keypoints $\mathcal{K}_{VR}(I_2)$ of image $I_2$.
	If we set $I_1 = I_M$ and $I_2 = I_D$ this gives us an approximation that accounts for the perspective change and the deformations caused by the HMD.

	\subsubsection{Mouth Camera Emulation during Training}
	Unfortunately, the alignment problem of mouth camera images and entire faces without an HMD makes it impossible to obtain perfect ground truth pairs for training.
    In our baseline approach~\cite{baseline}, the information bottleneck posed by the VR keypoints enables training on large-scale
	speaking head datasets which helps generalization to unseen persons without finetuning.
	
\begin{figure*}[htb!]
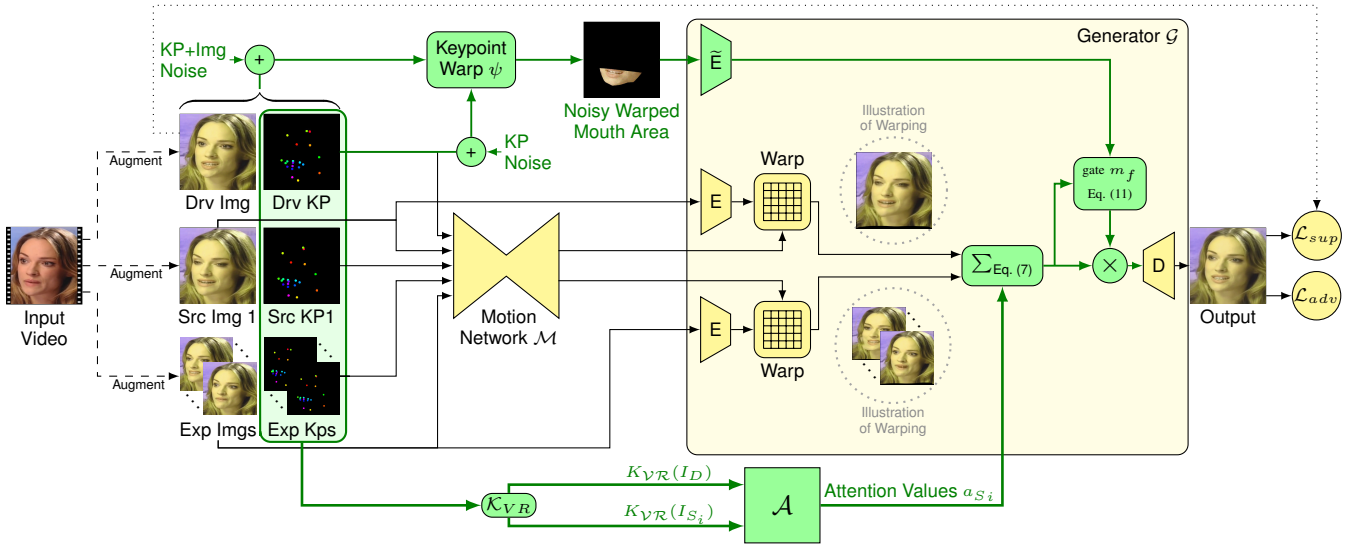

		\centering\newlength{\vidh}\setlength{\vidh}{.45cm}
		\begin{tikzpicture}[
		font=\sffamily\scriptsize,
		align=left,
		inp/.style={draw,minimum width=1cm,minimum height=1cm,inner sep=0},
		inp_c/.style={draw, circle, dotted, gray!70, thick, minimum width=1cm,minimum height=1cm,inner sep=0},
		inp_sm/.style={draw=none,minimum width=0.7cm,minimum height=0.7cm,inner sep=0}
		inp_nb/.style={draw=none,minimum width=1cm,minimum height=1cm,inner sep=0},
		m/.style={draw,rounded corners,fill=yellow!50,align=center},
		node distance=0.6,
		every label/.style={font=\sffamily\scriptsize,align=center,inner sep=2pt,text depth=1pt},
		label distance=0pt,
		]
		\node[inp,label=south:Drv Img] (driv_img) {\includegraphics[width=1cm]{images_2023/train_fig/id10037_CJ1wW4h9ZGM_000355_000690/driving.png}};
		\node[inp,right=0.1 of driv_img,label=south:Drv KP] (driv_kp) {\includegraphics[width=1cm]{images_2023/train_fig/id10037_CJ1wW4h9ZGM_000355_000690/kp_driving_s4.png}};
		
		\node[inp,below=.5 of driv_img, label=south:Src Img 1] (source_img) {\includegraphics[width=1cm]{images_2023/train_fig/id10037_CJ1wW4h9ZGM_000355_000690/source_0.png}};
		\node[inp,right=0.1 of source_img,label=south:Src KP1] (source_kp) {\includegraphics[width=1cm]{images_2023/train_fig/id10037_CJ1wW4h9ZGM_000355_000690/kp_source_s4_0.png}};

		\node[inp_sm] at ($(source_img.south)+(-0.15,-0.8)$) (expr_img) {\includegraphics[width=0.7cm]{images_2023/train_fig/id10037_CJ1wW4h9ZGM_000355_000690/source_1.png}};
		\node[inp_sm] at ($(expr_img.north)+(0.31,-0.8)$) (expr_img2) {\includegraphics[width=0.7cm]{images_2023/train_fig/id10037_CJ1wW4h9ZGM_000355_000690/source_2.png}};
        \node[right=0.2,inner sep=1pt,rotate=-50] at ($(expr_img.north)+(0.15,-0.15)$) (dots1) {\dots};
        \node[right=0.2,inner sep=1pt,rotate=-50] at ($(expr_img.south)+(-0.5,0.12)$) (dots2) {\dots};
        \node[fit=(expr_img)(expr_img2),draw=none,label={[yshift=0.3em]south:Exp Imgs},inner sep=0] (wrapper_expres_img) {};

		\node[inp_sm,right=0.64 of expr_img] at ($(source_img.south)+(-0.15,-0.8)$) (expr_kp) {\includegraphics[width=0.7cm]{images_2023/train_fig/id10037_CJ1wW4h9ZGM_000355_000690/kp_source_s4_1.png}};
		\node[inp_sm] at ($(expr_kp.north)+(0.31,-0.8)$) (expr_kp2) {\includegraphics[width=0.7cm]{images_2023/train_fig/id10037_CJ1wW4h9ZGM_000355_000690/kp_source_s4_2.png}};
        \node[right=0.2,inner sep=1pt,rotate=-50] at ($(expr_kp.north)+(0.15,-0.15)$) (dots3) {\dots};
        \node[right=0.2,inner sep=1pt,rotate=-50] at ($(expr_kp.south)+(-0.5,0.12)$) (dots4) {\dots};
        \node[fit=(expr_kp)(expr_kp2),draw=none,label={[yshift=0.3em]south:Exp Kps},inner sep=0] (wrapper_expres_kp) {};

		\node[inp,label=south:Input\\Video] (inb) at ($(source_img.west)+(-1.8,0)$) {\includegraphics[width=1cm]{images_2023/train_fig/id10037_CJ1wW4h9ZGM_000355_000690/video_img.png}};
		
		\node[inner sep=0,right=1.5 of source_kp,label={[yshift=0.6em]south:Motion\\Network $\mathcal{M}$}] (motion) {\tikz[scale=0.7]{
				\draw[fill=yellow!50] (-1,-1) -- (-1,1) -- (0,0.2) -- (1,1) -- (1,-1) -- (0,-0.2) -- cycle;}
		};

		\node[draw,fill=yellow!50,trapezium,shape border rotate=270] at ($(motion.east)+(2.0,-0.85)$) (expr_enc) {E};

		\node[m,anchor=west,minimum height=0.5cm,label=south:Warp] at ($(expr_enc.east)+(0.3,0)$) (warp_expr) {\tikz{
		\draw[step=0.1,black,thin] (0.0,0.0) grid (0.5,0.5);
        }};
    
		\node[inp_sm] at ($(warp_expr.east)+(+0.9,-0.05)$) (deformed_ex1) {\includegraphics[width=0.7cm]{images_2023/train_fig/id10037_CJ1wW4h9ZGM_000355_000690/deformed_1.png}};
		\node[inp_sm] at ($(deformed_ex1.north)+(0.38,-0.8)$) (deformed_ex2) {\includegraphics[width=0.7cm]{images_2023/train_fig/id10037_CJ1wW4h9ZGM_000355_000690/deformed_2.png}};
		\node[right=0.2,inner sep=1pt,rotate=-50] at ($(deformed_ex1.north)+(0.15,-0.15)$) (dots1) {\dots};
		\node[right=0.2,inner sep=1pt,rotate=-50] at ($(deformed_ex1.south)+(-0.44,0.12)$) (dots2) {\dots};
		\node[draw,dotted,circle, inner sep=-3.1, gray!70, thick, fit=(deformed_ex1)(deformed_ex2),label={[name=label_deformed_expr,yshift=0.1em]south:\textcolor{gray!90}{\scalefont{0.75}Illustration}\\[-0.07cm] \textcolor{gray!90}{\scalefont{0.75}of Warping}}] (deformed_expr) {};

		\node[draw,fill=yellow!50,trapezium,shape border rotate=270,anchor=west] at ($(motion.east)+(1.8,0.85)$) (source_enc) {E};
		
		\node[m,anchor=west,minimum height=0.5cm,label=north:Warp] at (source_enc-|warp_expr.west) (warp_source) {\tikz{
		\draw[step=0.1,black,thin] (0.0,0.0) grid (0.5,0.5);
}};

		\node[inp_c,label={[yshift=-0.1em]north:\textcolor{gray!90}{\scalefont{0.75}Illustration}\\[-0.07cm] \textcolor{gray!90}{\scalefont{0.75}of Warping}}]  at ($(warp_source.east)+(1.1,0.15)$) (deformed_source) {{\includegraphics[width=1cm]{images_2023/train_fig/id10037_CJ1wW4h9ZGM_000355_000690/deformed_0.png}}};

		\node[m,fill=green!40] at ($(motion.north)+(-0.5,2.025)$) (delauny_transform) {Keypoint\\Warp $\mathbf{\psi}$};
		\draw[black,decorate,decoration={calligraphic brace,amplitude=6pt}, thick] ($(driv_img.north west)+(0pt,1.9pt)$)--($(driv_kp.north east)+(0pt,1.9pt)$) node[midway, above](brace){};
		
		\node[circle,draw,fill=green!40, inner sep = 2pt] at (brace |- delauny_transform) (noise_delauny1) {+};
		\node[circle,draw,fill=green!40, inner sep = 2pt] at (driv_img -| delauny_transform) (noise_delauny2) {+};
		
        \node[inp,label=south:  \textcolor{green!50!black}{Noisy Warped}\\ \textcolor{green!50!black}{Mouth Area},fill=green!40] at ($(delauny_transform)+(2.0,0)$) (delauny_driv) {{\includegraphics[width=1cm]{images_2023/train_fig/id10037_CJ1wW4h9ZGM_000355_000690/delauny.png}}};

		\node[m,anchor=west,fill=green!40] at ($(warp_expr.east)!0.5!(warp_source.east)+(2.0,0)$) (sum) {$\sum_{\text{\cref{eq:attention}}}$};
		\node[m, circle, fill=green!40,font=\normalsize,inner sep=1pt,right=0.65cm of sum] (odot) {$\times$};
		\node[m,font = \normalsize,above=0.5cm of odot,fill=green!40] (gating) {\tiny{gate $m_f$} \\ [-0.1cm] \tiny \cref{eq:gating}};
		\node[draw,fill=yellow!50,trapezium,shape border rotate=90,right=0.2 of odot] (dec) {D};
		\node[inp,right=0.2 of dec,label=south:Output] (out) {{\includegraphics[width=1cm]{images_2023/train_fig/id10037_CJ1wW4h9ZGM_000355_000690/out.png}}};
		\node[draw,circle,fill=yellow!50,right=0.3 of out,inner sep=0pt] at($(out.east)+(0.04,0.4)$) (loss_sub) {$\mathcal{L}_{sup}$};
		\node[draw,circle,fill=yellow!50,right=0.3 of out,inner sep=0pt] at($(out.east)+(0.04,-0.4)$) (loss_adv) {$\mathcal{L}_{adv}$};
		\node[draw,fill=green!40,trapezium,shape border rotate=270] at (source_enc |- delauny_driv) (mouth_enc) {$\widetilde{\text{E}}$};

		\begin{pgfonlayer}{background}
		\node[fit=(mouth_enc)(sum)(expr_enc)(source_enc)(dec)(label_deformed_expr),draw=black,rounded corners,inner sep=5pt,fill= yellow!12] (g) {};
		\node[anchor=north east,align=right] at (g.north east) {Generator $\mathcal{G}$};
		\end{pgfonlayer}
		\node[inp,fill=green!40,font=\normalsize,shape border rotate=90,below=1.45 of warp_expr] (attention) {$\mathcal{A}$};
		\node[m,fill=green!40, inner sep = 1.6pt] (extract_vr_kp) at (motion |- attention) {$\mathcal{K}_{VR}$};

		\begin{scope}[-latex]
		\draw[dashed] (inb.east) -- ++(0.15,0) |- (source_img) coordinate[pos=1.05] (augmenta);
		\draw[dashed] (inb.east)++(0,0.35) -- ++(0.2,0) |- (driv_img.west) coordinate[pos=1.05] (augmentb);
		\draw[dashed] (inb.east)++(0,-0.35) -- ++(0.2,0) |- ($(wrapper_expres_img.west)+(0.12,0)$) coordinate[pos=1.0] (augmentc);
		
		\draw (source_img.north) -- ++(0,0.1) -| ($(source_kp.east)!0.5!(motion.west)+(0,+0.8)$) |- ($(source_enc.west)+(0,0)$);
		\draw (source_img.north) -- ++(0,0.1) -| ($(source_kp.east)!0.5!(motion.west)+(0,+0.2)$) -- ($(motion.west)+(0,0.2)$);
		\draw (source_kp.east) -- ($(motion.west)$);
		\draw (driv_kp) -| ($(motion.west)+(-0.2,0.4)$) -- ($(motion.west)+(0,0.4)$);
		
		\draw (wrapper_expres_kp.east) -- ++(-0.2,0) -| ($(source_kp.east)!0.5!(motion.west)+(0,-0.2)$) -- ($(motion.west)+(0,-0.2)$);
		\draw (wrapper_expres_img.south) ++(0,-0.21) -- ++(0,-0.1) -| ($(expr_enc.west)+(-1.2,0.0)$) -- ($(expr_enc.west)+(0,0)$);
		\draw (wrapper_expres_img.south) ++(0,-0.21) -- ++(0,-0.1) -| ($(motion.west)+(-0.2,-0.4)$)
		-- ($(motion.west)+(0,-0.4)$);
		
		\draw ($(motion.east)+(-0.09,0.2)$) -| (warp_source.south) ;
		\draw (source_enc) --(warp_source);
		\draw (warp_source.east) --++ (0.1,0)|- ($(sum.west)+(0,+0.15)$); 
		
		\draw ($(motion.east)+(-0.09,-0.2)$) -| (warp_expr.north);
		\draw (expr_enc) -- (warp_expr);
		
		\draw (warp_expr.east) --++ (0.1,0)|- ($(sum.west)+(0,-0.15)$); 
		
		\draw[black!50!green, line width=0.25mm] (sum) -- (odot);
		\draw[black!50!green, line width=0.15mm] (odot) -- (dec);
        \draw[black!50!green, line width=0.25mm] ($(sum.east)+(0.15,0.0)$) |- (gating.west);
        \draw[black!50!green, line width=0.25mm] (mouth_enc) -| (gating);
        \draw[black!50!green, line width=0.25mm] (gating) -- (odot);
		\draw (dec) -- (out);
 		\draw (out.east)++(0,0.4) -- (loss_sub);
 		\draw (out.east)++(0,-0.4) -- (loss_adv);
		
		\draw [black!50!green, line width=0.35mm](attention) -| (sum) node [black,pos=0.245,above,inner sep=1pt,black!50!green] {Attention Values $\medmath{{a_S}_i}$};
		\draw [black!50!green, line width=0.35mm] ($(wrapper_expres_kp.south)+(0,-0.24)$) |- (extract_vr_kp);
		\draw [black!50!green, line width=0.35mm] (extract_vr_kp.north) |- ($(attention.west)+(0,+0.275)$) node [black,pos=0.84,above,inner sep=1pt,black!50!green] (tag1) {$\medmath{K_{\mathcal{VR}}(I_{D})}$};;
		\draw [black!50!green, line width=0.35mm] (extract_vr_kp.south) |- ($(attention.west)+(0,-0.275)$) node [black,inner sep=1pt,black!50!green] at ($(tag1.south)+(0,-0.39)$) {$\medmath{K_{\mathcal{VR}}(I_{S_i})}$};;

		\draw[-,black!50!green, line width=0.25mm] (brace.north) -- (noise_delauny1.south);
		\draw[black!50!green, line width=0.25mm] (noise_delauny1.east) -- ($(delauny_transform.west)$);
		\draw[-,black!50!green, line width=0.25mm] (driv_kp.east) -- (noise_delauny2);
		\draw[black!50!green, line width=0.25mm] (noise_delauny2.north) -- (delauny_transform);
		\draw[black!50!green, line width=0.25mm] (delauny_transform) -- (delauny_driv);
		\draw[black!50!green, line width=0.25mm] (delauny_driv) -- (mouth_enc);
		
		\draw[dotted] (driv_img.west)++(0,0.25) -- ++ (-0.35,0) --++ (0,1.7) -| (loss_sub.north);
		
		\end{scope}
		
		\node[left=0.2 of noise_delauny1,inner sep=1pt] (noise_in_delauny1) {\textcolor{green!50!black}{KP+Img}\\\textcolor{green!50!black}{Noise}};
		\draw[-latex,black!50!green] (noise_in_delauny1) -- (noise_delauny1);
		
		\node[right=0.2 of noise_delauny2,inner sep=1pt] (noise_in_delauny2) {\textcolor{green!50!black}{KP}\\\textcolor{green!50!black}{Noise}};
		\draw[-latex,black!50!green] (noise_in_delauny2) -- (noise_delauny2);

        \begin{pgfonlayer}{background}
		\draw[fill=green!10,rounded corners,draw=black!50!green, thick]
		($(driv_kp.north west)+(-0.05,0.05)$) -|
		($(wrapper_expres_kp.south east)+(-0.05,-0.23)$) -|
		($(wrapper_expres_kp.south west)+(+0.08,-0.23)$)-- cycle;
		\end{pgfonlayer}

		\node[font=\sffamily\tiny,anchor=north east,inner sep=1pt] at ($(augmenta)+(-0.3,0)$) {Augment};
		\node[font=\sffamily\tiny,anchor=north east,inner sep=1pt] at ($(augmentb)+(-0.3,0)$) {Augment};
		\node[font=\sffamily\tiny,anchor=north east,inner sep=1pt] at ($(augmentc)+(-0.15,0)$) {Augment};

		\end{tikzpicture} \vspace{-4.5ex}
		\caption{Training the facial animation network from videos. New components compared to our previous work~\cite{baseline} are highlighted in green.
			The supervised training loss $\mathcal{L}_{sup}$ is minimized when the network reconstructs the driving image given the source images, source keypoints, noisy warped mouth area, and the driving keypoints. Furthermore, a keypoint-aware discriminator network judges the quality of the generated image ($\mathcal{L}_{adv}$). The source images are chosen randomly from the input video. During training, we simulate mouth camera guidance by utilizing the lower facial area of the driving image itself. We regularize the mouth camera guidance by injecting image noise and keypoint noise. This simulates different lighting and imperfect keypoint warping as present during inference when the real mouth camera image is utilized.} \vspace{-2.5ex}
		\label{fig:train} 
	\end{figure*}

	To maintain this behavior and still provide visual mouth camera information, we propose a training-time data augmentation scheme. We add different types of camera noise \cite{noise}, but also simulate imperfect transformation by performing a keypoint warping on the driving frame to itself ($I_1 = I_2 = I_D$) with noise added to the keypoints (see \cref{fig:train}).
	The noise-augmented keypoint sequences are given by augmenting with (i) a normal distributed random scaling factor of the keypoint vector, (ii) a normal distributed random translation of the keypoint vector, and (iii) a normal distributed offset for each keypoint in the vector.

	During training, the resulting keypoint warping function 	
	\begin{align}
	 \psi(\,\,\omega^I[\,I_D\,],\omega^{K}[\,\medmath{\mathcal{K}_{VR}(I_D)}\,],\omega^{K}[\,\medmath{\mathcal{K}_{VR}(I_D)}\,]\,\,) 
	\end{align} therefore only utilizes $I_D$ in combination with an image noise operator $\omega^{I}$ and a keypoint noise operator $\omega^{K}$ (see \cref{fig:train}).

    \subsubsection{Gating Network}
    Additionally, we allow usage of the warped mouth area only through gated convolutions, which prevents
	direct information propagation. 
    We feed the keypoint-warped representation of the mouth area into the mouth image encoder $\widetilde{E}$ (see generator in \cref{fig:infer,fig:train}) that has a downsampling factor of four. This estimates the warped mouth area features 
    \begin{align}
    \widetilde{E}_M =&\,\,\widetilde{E}(\,\, \psi(\,I,\medmath{\mathcal{K}_{VR}(I)},\medmath{\mathcal{K}_{VR}(I_D)}\,)\,\,),   \label{eq:E_M}
    \end{align}
    where $\psi(\cdot)$ performs the keypoint warping from image $I$ into the driving VR keypoints $\mathcal{K}_{VR}(I_D)$.

    For gating the aggregated deformed source features $\E{S}{D}$, we concatenate the warped mouth area features $\widetilde{E}_M$ with $\E{S}{D}$ 
    and feed them through a small residual network with two layers to compute the gating weights (see \cref{fig:infer}).
    The resulting features in the main branch are therefore given by
    
    \begin{align}
      f &= \underbrace{\sigma(\,\,\phi[ \widetilde{E}_M  \oplus \E{S}{D}]\,\,)}_{=:m_f} \odot \E{S}{D},  \label{eq:gating}
    \end{align}

    where $\odot$ is the elementwise multiplication, $\oplus$ is concatenation, $\sigma(\cdot)$ is the sigmoid function, and $\phi[\cdot]$ is the convolutional feature extraction of the small residual network.
    
    Inducing visual mouth camera information implicitly through gating allows to mask out incorrect activations in the aggregated deformed source features $\E{S}{D}$ (see \cref{eq:attention}) while still being able to encode additional information without direct information propagation. This is especially beneficial when performing inter-operator animation (see \cref{fig:q_fancy}) or generalizing from entire faces during training to mouth camera images during inference.

	\subsection{Training} \label{sec:train}
	The training pipeline is visualized in \cref{fig:train}.
	All modules are trained end-to-end on the speaking head dataset Vox-Celeb~\citep{vox}.
	We train with perceptual loss and utilize a keypoint-aware discriminator network to generate adversarial losses, similar to \citet{fom}.
	Given a video, we randomly choose one driving frame and $n$$\,=\,$$5$ different source images, from which the last four are expressional source images.
	We extract facial keypoints and estimate a deformation grid of all source images into the driving keypoints using the motion network $\mathcal{M}$.
	Simultaneously, we estimate the attention values $a_{S_i}$.
	The source image features are then deformed and aggregated in the lower facial area using the attention values (see \cref{eq:attention}). The features are conditioned with the keypoint warped mouth area (see \cref{eq:gating}) and decoded to the output image.

	We initialize the keypoint detector, motion network, generator-encoder, and generator-decoder with weights of our baseline.
	The new components (attention mechanism $\mathcal{A}$ and gating network) are trained from scratch.
	We found that the initialization with the baseline weights resulted in a very fast progress.

	\textit{Finetuning with Imperfect VR Annotations: }
	Unlike the baseline, our proposed method allows to explicitly train with mouth camera images.
	We therefore extend the datasets with some VR facial animation samples. We annotate such samples by manually searching for correspondences in the mouth camera and the face camera. The manual alignment is a very challenging task and often there is no perfect solution.
	Due to time limitations and efficiency reasons, we only annotate 13 different operators of our system and chose a fraction of training samples from such imperfect annotations.
	To prevent overfitting, we randomly scale, rotate and crop the facial images.
	Random cropping followed by rescaling to a quadratic image also changes face aspect ratio.

    During finetuning, we select 6\% of the training samples from the annotated VR datasets and 94 \% from the Vox-Celeb dataset, which gives similar importance to our annotated videos and videos from Vox-Celeb.
    
	\subsection{Inference}
	Preprocessing, which is explained in \cref{sec:baseline}, remains equivalent to our baseline and takes approximately 15 minutes.
	We then select $n$$=$$4$ or $n$$=$$5$ fixed source images with different facial expressions.
	Given the current mouth camera image we optionally retrieve the best matching image (see \cref{sec:baseline}) which will be treated as an expressional source image.
	Following our baseline, we then construct the imaginary driving keypoints using the mapped mouth camera keypoints, the 3D head pose keypoints from the first source image, and the eye tracking results.
	The deformation, attention and refinement steps are equivalent to the training pipeline.
	During inference, the keypoint warping and gating step is always performed with the mouth camera image from the HMD. We therefore set $\widetilde{E}_M$ (see \cref{eq:E_M}) in \cref{eq:gating} to
    \begin{align}
      \widetilde{E}_M = \widetilde{E}(\,\, \psi(\,I_M,\medmath{\mathcal{K}_{VR}(I_M)},\Pi_S(\medmath{\mathcal{K}_{VR}(I_M)})\,)\,\,), \label{eq:E_M_infer}
    \end{align}
    where $I_M$ is the mouth camera image and $\Pi_S(\medmath{\mathcal{K}_{VR}(I_M)})$ are the mouth camera keypoints projected into the first source image (the VR keypoints of the imaginary driving frame).
    
	\subsection{Temporal Consistency} \label{sec:tc} The baseline often struggles generating temporally consistent facial animations.
	The abrupt change of the expression frame induces the greatest negative influence.
	
	Our proposed attention mechanism can be used in two different configurations. 
	In the first configuration, all $n$$=$$5$ source frames are fixed, which minimizes the temporal inconsistencies as the continuous attention weight function changes smoothly with the mouth camera stream.
	The second configuration allows to retrieve the last expressional source image $I_{S_5}$ during inference dynamically, which improves the output quality slightly (see \cref{tab:eval:ablations}).
	To control the risk of temporal inconsistencies, we introduce a maximum attention value $a_{max}$ for the retrieved images in the attention mechanism. 
	This parameter allows us to control the tradeoff between quality and temporal consistency (see \cref{tab:eval:tc}).
	During testing, we set $a_{max}$ operator-specifically but with a default value of $25$\%.
	In case the image retrieval does not perform well, the $a_{max}$ value can be reduced.

	Furthermore, the proposed visual mouth camera guidance reduces the network's dependency on the retrieved image which also contributes to the temporal consistency.

	\begin {table*}[htb!]
    \centering \footnotesize
    \setlength{\tabcolsep}{2pt}
    \begin{threeparttable}
    \caption{Ablation study}
	\label{tab:eval:ablations}
    \begin{tabular}{l@{\hspace{8pt}}ccc@{\hspace{7pt}}|@{\hspace{7pt}}ccc@{\hspace{14pt}}ccc@{\hspace{14pt}}ccc@{\hspace{14pt}}ccc@{\hspace{14pt}}ccc} \toprule

        &  \multicolumn{3}{c@{\hspace{14pt}}}{\textbf{MEAN}} & \multicolumn{3}{c@{\hspace{14pt}}}{Male1}& \multicolumn{3}{c@{\hspace{14pt}}}{Male2}& \multicolumn{3}{c@{\hspace{14pt}}}{Male3} & \multicolumn{3}{c@{\hspace{14pt}}}{Fem1} & \multicolumn{3}{@{}c@{}@{}}{Fem2}

        \\ \cmidrule (r{14pt}) {5-7}
        \cmidrule (r{14pt}) {2-4}
        \cmidrule (r{14pt}) {8-10}
        \cmidrule (r{14pt}) {11-13}
        \cmidrule (r{14pt}) {14-16}
        \cmidrule (r{3pt}) {17-19}
        Method       & psnr    & ssim     &lpips   & psnr    & ssim     &lpips   & psnr    & ssim     &lpips  & psnr    & ssim     &lpips  & psnr    & ssim     &lpips  & psnr    & ssim     &lpips \\ \midrule
        Ours-50\%    & \textbf{28.83}  & \textbf{.8603}   & \textbf{.0357} & \textbf{29.27}  & \textbf{.8642}   & \textbf{.0365} & \textbf{28.66} & \textbf{.8610}   & .0368  & \textbf{28.59}  & \textbf{.8439}   & .0368 & \textbf{29.87}  & \textbf{.9028}   & \textbf{.0233}& \textbf{27.74}  & \textbf{.8298}   & \textbf{.0451} \\
        Ours         & 28.75   & .8586    & .0361  & 29.08   & .8603    & .0373  & 28.63   & .8602    & .0370  & 28.45   & .8410    & .0375 & 29.87   & .9023    & .0235 & 27.72   & .8294    & .0452 \\
        Ours-5-Fix   & 28.50   & .8504    & .0376  & 28.87   & .8494    & .0401  & 28.30   & .8521    & .0383  & 28.06   & .8274    & .0399 & 29.66   & .8974    & .0238 & 27.59   & .8257    & .0461 \\ 
        Ours-Short   & 28.28   & .8550    & .0376  & 28.64   & .8486    & .0437  & 28.45   & .8589    & \textbf{.0318} & 28.19   & .8436    & \textbf{.0364}& 28.69   & .8963    & .0246 & 27.42   & .8275    & .0515 \\
        Ours-NF      & 27.20   & .8369    & .0465  & 27.77   & .8350    & .0432  & 27.36   & .8363    & .0467  & 27.13   & .8267    & .0437 & 27.50   & .8842    & .0357 & 26.23   & .8024    & .0630 \\
        Base~\citep{baseline}& 25.10   & .7809    & .0580  & 24.68   & .7513    & .0646  & 24.97   & .7974    & .0585  & 26.79   & .7868    & .0470 & 23.89   & .8108    & .0472 & 25.19   & .7584    & .0728 \\
        \midrule
        
        Ours-10-Skip & 28.69   & .8568    & .0363  & 29.03   & .8582    & .0372  & 28.52   & .8566    & .0375  & 28.40   & .8399    & .0380 & 29.80   & .9010    & .0236 & 27.71   & .8283    & .0452 \\ 
        Base-10-Skip & 24.96   & .7758    & .0596  & 24.61   & .7469    & .0653  & 24.91   & .7902    & .0589  & 26.69   & .7858    & .0481 & 23.78   & .8074    & .0500 & 24.83   & .7486    & .0756 \\ %

        \bottomrule
    \end{tabular} %
     \textit{NF}: No finetuning on mouth camera images, \textit{Short}: finetuning for a short time which leads to only 4000~VR~images in the training batches,
     \textit{10-Skip}: only one out of ten source video images retrievable,
     \textit{Ours}: maximum image retrieval attention parameter $a_{max}$$\,=\,$$25\%$,
     \textit{Ours-50\%}: $a_{max}$$\,=\,$$50\%$,
     \textit{Ours-5-Fix}: only five fixed source images without image retrieval.
	\end{threeparttable} \vspace{-3ex}
    \end{table*}

	\section{Experiments and Evaluation}
	We compare against our baseline method~\cite{baseline} which we used at the ANA Avatar XPRIZE Semifinals.
	A fair comparison to other methods \cite{lombardi2018deep,vr_facial,faceaudio} is not feasible as they perform per operator optimization with a significant amount of training, preprocessing, and data capturing.
	All reported qualitative and quantitative results are obtained with unseen persons.

\begin{figure}[h]
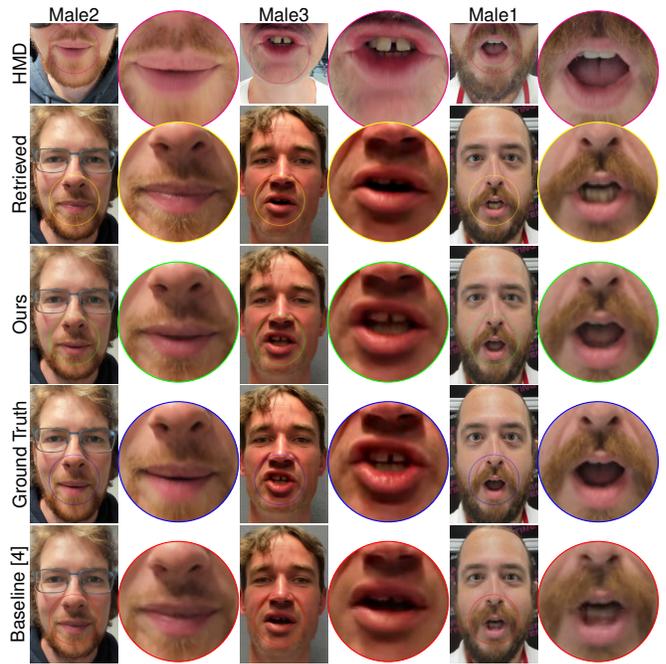

  \centering
  \begin{tikzpicture}[font=\sffamily\scriptsize,spy using outlines={circle, magnification=2.4, size=0.09\textwidth,every spy in node/.append style={thin,opacity=1.0}}]
    \node[inner sep=0,matrix, matrix of nodes, nodes={inner sep=0}, row sep=1pt, column sep=46.5pt,    
    column 1/.style={anchor=center, column sep=1pt},
    column 2/.style={anchor= west},
    column 3/.style={anchor= east},
    column 4/.style={anchor=center}] (m) {
      |[rotate=90]| HMD &
      |(in1)|\includegraphics[width=0.065\textwidth,clip,trim=3.5cm 0.6cm 3.5cm 0cm]{images_2023/vgl_gt/chris/66__mouth_scaled4.png} &
      |(in2)|\includegraphics[width=0.065\textwidth,clip,trim=3.5cm 0.6cm 3.5cm 0cm]{images_2023/vgl_gt/mischa/77__mouth_scale4.png} &
      |(in3)|\includegraphics[width=0.065\textwidth,clip,trim=3.5cm 0.6cm 3.5cm 0cm]{images_2023/vgl_gt/judge2/57__mouth_scale4.png} \\
      |[rotate=90]| Retrieved &
		|(ret1)|\includegraphics[width=0.065\textwidth,clip,trim=1.75cm 0cm 1.75cm 0.3cm]{images_2023/vgl_gt/chris/66_express.png} &
		|(ret2)|\includegraphics[width=0.065\textwidth,clip,trim=1.75cm 0cm 1.75cm 0.3cm]{images_2023/vgl_gt/mischa/77_express.png} &
		|(ret3)|\includegraphics[width=0.065\textwidth,clip,trim=1.75cm 0cm 1.75cm 0.3cm]{images_2023/vgl_gt/judge2/57_express.png} \\
      |[rotate=90]| Ours &
      |(ours1)|\includegraphics[width=0.065\textwidth,clip,trim=1.75cm 0cm 1.75cm 0.3cm]{images_2023/vgl_gt/chris/66_out.png} &
      |(ours2)|\includegraphics[width=0.065\textwidth,clip,trim=1.75cm 0cm 1.75cm 0.3cm]{images_2023/vgl_gt/mischa/77_out.png} &
      |(ours3)|\includegraphics[width=0.065\textwidth,clip,trim=1.75cm 0cm 1.75cm 0.3cm]{images_2023/vgl_gt/judge2/57_out.png} \\
      |[rotate=90]| Ground Truth &
      |(gt1)|\includegraphics[width=0.065\textwidth,clip,trim=1.75cm 0cm 1.75cm 0.3cm]{images_2023/vgl_gt/chris/66_gt.png} &
      |(gt2)|\includegraphics[width=0.065\textwidth,clip,trim=1.75cm 0cm 1.75cm 0.3cm]{images_2023/vgl_gt/mischa/77_gt.png} &
      |(gt3)|\includegraphics[width=0.065\textwidth,clip,trim=1.75cm 0cm 1.75cm 0.3cm]{images_2023/vgl_gt/judge2/57_gt.png} \\
      |[rotate=90]| Baseline~\cite{baseline} &
      |(base1)|\includegraphics[width=0.065\textwidth,clip,trim=1.75cm 0cm 1.75cm 0.3cm]{images_2023/vgl_gt/chris/66_out_base.png} &
      |(base2)|\includegraphics[width=0.065\textwidth,clip,trim=1.75cm 0cm 1.75cm 0.3cm]{images_2023/vgl_gt/mischa/77_out_base.png} &
      |(base3)|\includegraphics[width=0.065\textwidth,clip,trim=1.75cm 0cm 1.75cm 0.3cm]{images_2023/vgl_gt/judge2/57_out_base.png} \\
    };
     \coordinate (a) at ($(in1.north)+(0,-0.1)$);
     \node[above=.0 of a] {Male2};
     
     \coordinate (a) at ($(in2.north)+(0,-0.1)$);
     \node[above=.0 of a] {Male3};
     
     \coordinate (a) at ($(in3.north)+(0,-0.1)$);
     \node[above=.0 of a] {Male1};

    \spy [magenta, opacity=0.35,every spy on node/.append style={ultra thin}] on (rel cs:x=50,y=68.5,name=in1) in node [right] at ($(in1.east)+(0,-0.1)$);
	\spy [magenta, opacity=0.35, magnification=2,every spy on node/.append style={ultra thin}] on (rel cs:x=45,y=62.5,name=in2) in node [right] at ($(in2.east)+(0,-0.1)$);
	\spy [magenta, opacity=0.35,every spy on node/.append style={ultra thin}] on (rel cs:x=50,y=68.5,name=in3) in node [right] at ($(in3.east)+(0,-0.1)$);
	
    \spy [yellow, opacity=0.35,every spy on node/.append style={ultra thin}] on (rel cs:x=50,y=31.5,name=ret1) in node [right] at ($(ret1.east)+(0,-0.1)$);
	\spy [yellow, opacity=0.35,every spy on node/.append style={ultra thin}] on (rel cs:x=45,y=31.5,name=ret2) in node [right] at ($(ret2.east)+(0,-0.1)$);
	\spy [yellow, opacity=0.35,every spy on node/.append style={ultra thin}] on (rel cs:x=50,y=31.5,name=ret3) in node [right] at ($(ret3.east)+(0,-0.1)$);

    \spy [green, opacity=0.3,every spy on node/.append style={ultra thin}] on (rel cs:x=50,y=31.5,name=ours1) in node [right] at ($(ours1.east)+(0,-0.1)$);
    \spy [green, opacity=0.3,every spy on node/.append style={ultra thin}] on (rel cs:x=45,y=31.5,name=ours2) in node [right] at ($(ours2.east)+(0,-0.1)$);
    \spy [green, opacity=0.3,every spy on node/.append style={ultra thin}] on (rel cs:x=50,y=31.5,name=ours3) in node [right] at ($(ours3.east)+(0,-0.1)$);
    
    \spy [blue, opacity=0.3,every spy on node/.append style={ultra thin}] on (rel cs:x=50,y=31.5,name=gt1) in node [right] at ($(gt1.east)+(0,-0.1)$);
    \spy [blue, opacity=0.3,every spy on node/.append style={ultra thin}] on (rel cs:x=45,y=31.5,name=gt2) in node [right] at ($(gt2.east)+(0,-0.1)$);
    \spy [blue, opacity=0.3,every spy on node/.append style={ultra thin}] on (rel cs:x=50,y=31.5,name=gt3) in node [right] at ($(gt3.east)+(0,-0.1)$);
    
    \spy [red, opacity=0.35,every spy on node/.append style={ultra thin}] on (rel cs:x=50,y=31.5,name=base1) in node [right] at ($(base1.east)+(0,-0.1)$);
    \spy [red, opacity=0.35,every spy on node/.append style={ultra thin}] on (rel cs:x=45,y=31.5,name=base2) in node [right] at ($(base2.east)+(0,-0.1)$);
    \spy [red, opacity=0.35,every spy on node/.append style={ultra thin}] on (rel cs:x=50,y=31.5,name=base3) in node [right] at ($(base3.east)+(0,-0.1)$);
    
  \end{tikzpicture} \vspace{-4.2ex}
	\caption{Visual results of our quantitative analysis in \cref{tab:eval:ablations}. For all examples the image retrieval (second row) was inaccurate, which led to poor results for the baseline~\citep{baseline} (bottom).
	Our method still generates good results.}
	\label{fig:q_gt} \vspace{-3.5ex}
\end{figure}

\subsection{Quantitative Results}
To generate quantitative results, we utilize the annotated VR dataset. The mouth camera image is the input and the corresponding facial image will be the driving frame.  We evaluate our method on five different persons.
As our method is intended to improve the facial animation in the mouth region, we only measure the metrics Peak signal-to-noise ratio (PSNR), Structural Similarity (SSIM), and Learned Perceptual Image Patch Similarity (LPIPS)~\cite{lpips} in the lower half of the face without background.

\subsubsection{Accuracy}\label{sec:accuracy}
\cref{tab:eval:ablations} shows that all proposed model variants significantly outperform the baseline \cite{baseline}.
All of our ablations, besides Ours-Short and Ours-NF, in \cref{tab:eval:ablations} are trained for 50 epochs with annotated VR samples as explained in \cref{sec:train}.
The Ours-NF ablation, however, was never trained with VR samples.
In this case, the missing regularizing influence results in overfitting after roughly five epochs, so we report the results at this training step.
Interestingly, Ours-NF already outperforms the baseline in all metrics significantly.
Ours-Short is only finetuned for 5 epochs which corresponds to just 4000 VR/Face image pairs that have been seen.
This is already enough to generate similar results as obtained with 50 epochs VR finetuning.

Ours-10-Skip and Base-10-Skip represent ablations where only one out of ten images in the source video is retrievable. This results in a larger gap between the driving image and retrievable source images. 
When reducing retrievable images and thus the number of presented facial expressions by factor ten, quality is only influenced slightly (see mean metrics of Ours vs. Ours-10-Skip in \cref{tab:eval:ablations}).

Our second method variation (Ours-5-Fix) further limits the number of different facial expressions presented to the network. It has only five fixed source images and therefore uses no image retrieval.
The results indicate that image retrieval is not essential in our method for achieving good animations.
Comparing all our method ablations shows that $a_{max}$$=\,$$50\%$ generates the highest accuracy, but results in reduced temporal consistency, compared to $a_{max}$$=\,$$25\%$ and Ours-5-Fix without image retrieval, as evaluated in \cref{tab:eval:tc}.

\subsubsection{Temporal Consistency} \label{sec:eval:tc}
In our proposed method and baseline~\cite{baseline} temporal inconsistencies mainly occur whenever a new expression frame is selected, which happens in roughly every second frame when speaking.
Measuring temporal consistency in animated facial images is a non-trivial task, especially when disoclusions and complex facial deformations occur. To reduce these effects, we use the motion network to deform the previous prediction into the current one. This allows comparison using perceptual similarity (LPIPS~\cite{lpips}), with the assumption that two consecutive frames exhibit only small expressional differences. Importantly, unintended discontinuous flicker effects lead to large errors in this metric.
Note that the proposed measure does not necessarily correlate with accuracy.

In \cref{tab:eval:tc}, we report temporal inconsistency for four different persons from \cref{tab:eval:ablations}, which are ordered with a descending image-retrieval quality from left to right. The best temporal consistency is obtained without image retrieval (Ours-5-Fix). When using image retrieval, the measured temporal consistency decreases with the maximum attention parameter $a_{max}$ (see \cref{sec:tc}). Together with \cref{tab:eval:ablations}, this highlights the temporal consistency vs. accuracy tradeoff, which is controllable through $a_{max}$. However, compared to the baseline, all of our model variants perform much better, which is due to the baseline's strong dependence on the retrieved image.
The discrepancy to our method gets larger with worsening image retrieval quality.

To increase temporal consistency, the baseline method (Base+TCF) explicitly minimizes this measuring scheme by recursively low-pass filtering the retrieved expression image using the deformations of the last expression frame and the last prediction, which is exactly what we measure.
However, even if the image retrieval works fine, this comes with the cost of a reduced image quality in the lower facial area.
Even though our ablations already achieve significantly better results than Base+TCF, we equip an additional ablation using $a_{max}=50\%$ with the same recursive filtering scheme (Ours-50\%+TCF) to allow a fairer comparison. 

\begin{figure}
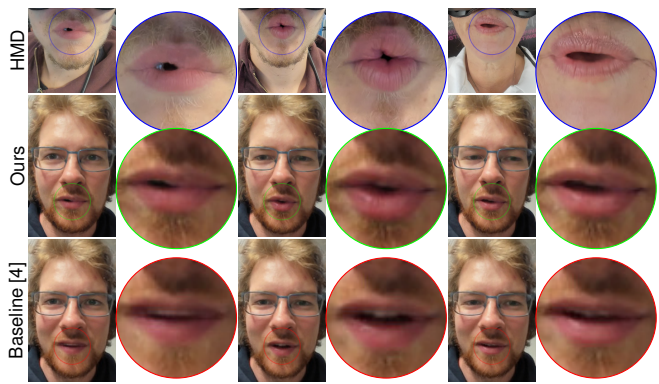

  \centering
  \begin{tikzpicture}[font=\sffamily\scriptsize, spy using outlines={circle, magnification=3.3, size=0.09\textwidth,every spy in node/.append style={thin,opacity=1}}]
    \node[inner sep=0,matrix, matrix of nodes, nodes={inner sep=0,anchor=center}, row sep=1pt, column sep=46.5pt,
    column 1/.style={anchor=center, column sep=1pt},
    column 2/.style={anchor= center},
    column 3/.style={anchor= center},
    column 4/.style={anchor=center}] (m) {
      |[rotate=90]| HMD &
      |(in1)|\includegraphics[width=0.065\textwidth,clip,trim=9cm 0cm 9cm 0cm]{images_2023/vgl_fancy/vr_238.png} &    
      |(in3)|\includegraphics[width=0.065\textwidth,clip,trim=9cm 0cm 9cm 0cm]{images_2023/vgl_fancy/vr_2704.png} &
      |(in5)|\includegraphics[width=0.065\textwidth,clip,trim=9cm 0cm 9cm 0cm]{images_2023/vgl_fancy/judge1_to_chris/vr_55.png}\\
      |[rotate=90]| Ours &
      |(out1)|\includegraphics[width=0.065\textwidth,clip,trim=1.75cm 0cm 1.75cm 0cm]{images_2023/vgl_fancy/out_238_eye_[-0.3, -0.5, 0].png} &
      |(out2)|\includegraphics[width=0.065\textwidth,clip,trim=1.75cm 0cm 1.75cm 0cm]{images_2023/vgl_fancy/out_2704_eye_[-0.3, 0.3, 0].png} &
      |(out3)|\includegraphics[width=0.065\textwidth,clip,trim=1.75cm 0cm 1.75cm 0cm]{images_2023/vgl_fancy/judge1_to_chris/out_55_eye_[0.3, 0.5, 0].png}\\
      |[rotate=90]| Baseline~\citep{baseline} &
      |(out1_b)|\includegraphics[width=0.065\textwidth,clip,trim=1.75cm 0cm 1.75cm 0cm]{images_2023/vgl_fancy/out_base_238_eye_[-0.3, -0.5, 0].png} &
      |(out2_b)|\includegraphics[width=0.065\textwidth,clip,trim=1.75cm 0cm 1.75cm 0cm]{images_2023/vgl_fancy/out_base_2704_eye_[-0.3, 0.3, 0].png} &
      |(out3_b)|\includegraphics[width=0.065\textwidth,clip,trim=1.75cm 0cm 1.75cm 0cm]{images_2023/vgl_fancy/judge1_to_chris/out_base_55_eye_[0.3, 0.5, 0].png} \\
    };
    \coordinate (a) at ($(in1.north)!0.5!(in2.north)$);

    \spy [red, opacity=0.35,every spy on node/.append style={ultra thin}] on (rel cs:x=50,y=25,name=out1_b) in node [right] at ($(out1_b.east)+(0,-0.1)$);
    \spy [blue,size=0.09\textwidth, magnification=2.6, opacity=0.3,every spy on node/.append style={ultra thin}] on (rel cs:x=50,y=72,name=in1) in node [right] at ($(out1.north east)+(0,0.3)$);
    \spy [green, opacity=0.3,every spy on node/.append style={ultra thin}] on (rel cs:x=50,y=25,name=out1) in node [right] at ($(out1.east)+(0,-0.3)$);
    
    \spy [red, opacity=0.35,every spy on node/.append style={ultra thin}] on (rel cs:x=50,y=25,name=out2_b) in node [right] at ($(out2_b.east)+(0,-0.1)$);
    \spy [blue,size=0.09\textwidth, magnification=2.6, opacity=0.3,every spy on node/.append style={ultra thin}] on (rel cs:x=50,y=72,name=in3) in node [right] at ($(out2.north east)+(0,0.3)$);
    \spy [green, opacity=0.3,every spy on node/.append style={ultra thin}] on (rel cs:x=50,y=25,name=out2) in node [right] at ($(out2.east)+(0,-0.3)$);
    
    \spy [red, opacity=0.3,every spy on node/.append style={ultra thin}] on (rel cs:x=50,y=25,name=out3_b) in node [right] at ($(out3_b.east)+(0,-0.1)$);
    \spy [blue,size=0.09\textwidth, magnification=2.6, opacity=0.35,every spy on node/.append style={ultra thin}] on (rel cs:x=50,y=72,name=in5) in node [right] at ($(out3.north east)+(0,0.3)$);
    \spy [green, opacity=0.3,every spy on node/.append style={ultra thin}] on (rel cs:x=50,y=25,name=out3) in node [right] at ($(out3.east)+(0,-0.3)$);

   \end{tikzpicture} \vspace{-3.5ex}
	\caption{VR facial animation from mouth camera input to the appearance of a different operator.
	Mouth camera guidance resolves keypoint ambiguities and models a broader range of mouth expressions (note the lips which partly stick together).}
	\label{fig:q_fancy} \vspace{-2ex}
\end{figure}

\begin{table}
\centering
\begin {threeparttable}
    \centering \footnotesize
    \setlength{\tabcolsep}{7pt}
    \caption{Temporal Inconsistency}
    \label{tab:eval:tc}
    \begin{tabular}{@{}lr@{.}lr@{.}lr@{.}lr@{.}l@{}} \toprule
Method        & \multicolumn{2}{c}{Male2}     & \multicolumn{2}{c}{Female1} & \multicolumn{2}{c}{Male3} & \multicolumn{2}{c}{Female2}  \\
\midrule
Ours-5-Fix       & \textbf{+0} & \textbf{0\,\%} & \textbf{+0} & \textbf{0\,\%} & \textbf{+0} & \textbf{0\,\%} & \textbf{+0} & \textbf{0\,\%} \\
Ours-25\%     &          +6 & 2\,\%          &          +5 & 9\,\%          &         +11 & 5\,\%          &          +8 & 7\,\% \\
Ours-50\%     &          +7 & 6\,\%          &          +8 & 3\,\%          &         +15 & 8\,\%          &         +16 & 8\,\% \\
Base~\citep{baseline}    &         +50 & 8\,\%          &         +86 & 2\,\%          &        +106 & 5\,\%          &        +151 & 9\,\% \\ \midrule 
Ours-50\%+TCF &         (+1 & 3\,\%)         &         (+1 & 3\,\%)         &         (+4 & 5\,\%)         &         (+4 & 4\,\%) \\
Base+TCF~\citep{baseline}&        (+21 & 6\,\%)         &        (+33 & 5\,\%)         &        (+45 & 0\,\%)         &        (+88 & 1\,\%) \\ 
\bottomrule
    \end{tabular} %
    Values normalized to Ours-5-Fix. Lower is better.
    25\% and 50\% indicate the $a_{max}$ parameter, TCF means temporal consistency filtering \citep{baseline}. Persons sorted by image retrieval quality (left: good).
\end {threeparttable}\vspace{-3.5ex}
\end{table}

\newlength{\eyeh}\setlength{\eyeh}{1.5cm}
\newcommand{\eyeplot}[1]{%
  \begin{axis}[
        ymin=-0.85,ymax=0.85,xmin=-0.85,xmax=0.85, scale only axis, width=\eyeh, height=\eyeh,
        axis equal,xtick=\empty,ytick=\empty,
        axis lines=center
   ]
   \addplot[red,only marks] table [x index=0, y expr=-\thisrowno{1},row sep=\\] {
        #1 \\
   };
  \end{axis}
}

\newcommand{\kpinput}[2]{%
  \tikz{
    \node[inner sep=0] (a) {\includegraphics[width=0.066\textwidth,clip,trim=13cm 0cm 13cm 0cm]{#1}};
    \node[inner sep=0, right=0.0 of a] {\tikz{\eyeplot{#2}}};
  }
}

\newcommand{\kpinputsm}[2]{%
  \tikz{
    \node[inner sep=0] (a) {\includegraphics[width=0.066\textwidth,clip,trim=9cm 0cm 9cm 0cm]{#1}};
    \node[inner sep=0, right=0.0 of a] {\tikz{\eyeplot{#2}}};
  }
}
\begin{figure*}[htb]
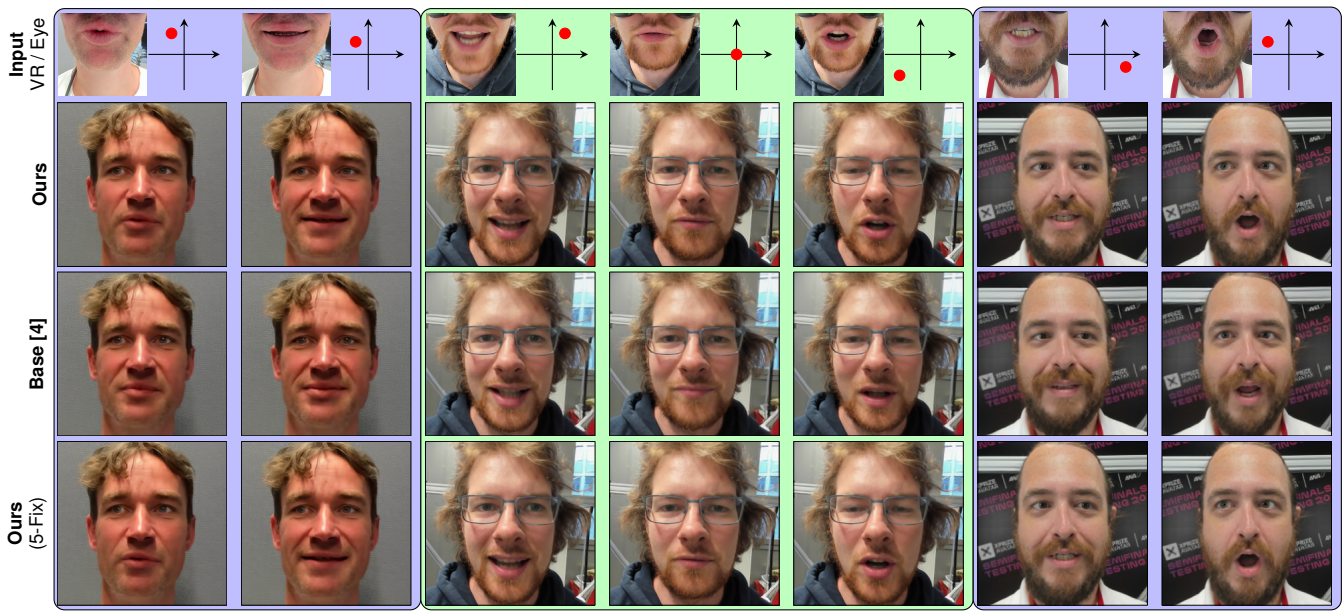
 \centering
	\setlength{\eyeh}{0.95cm}
	\tikz[font=\sffamily\scriptsize,align=center]{
      \node[matrix,matrix of nodes, nodes={inner sep=0,anchor=center},column sep=3.45pt,row sep=2pt] (m) {
		|(in1)| \kpinput{images_2023/vgl_infer/mischa_second_capture/vr_48.png}{-0.3 -0.5} &
		|(in2)|\kpinput{images_2023/vgl_infer/mischa_second_capture/vr_765.png}{-0.3 -0.3} &
		|(in3)|\kpinput{images_2023/vgl_infer/chris/vr_297.png}{0.3 -0.5} &
		|(in4)|\kpinput{images_2023/vgl_infer/chris/vr_255.png}{0 0} &
		|(in5)|\kpinput{images_2023/vgl_infer/chris/vr_173.png}{-0.5 0.5} &
		|(in6)|\kpinputsm{images_2023/vgl_infer/judge2/vr_303.png}{0.5 0.3} &
		|(in7)|\kpinputsm{images_2023/vgl_infer/judge2/vr_199.png}{-0.5 -0.3} & \\ [.0cm]
		|(out1)|\includegraphics[frame,width=0.127\textwidth,clip, trim=0cm 0cm 0cm 0.3cm]{images_2023/vgl_infer/mischa_second_capture/ours25/out_48_eye_[0.26344025, -0.17296118, 0].png} &
		\includegraphics[frame,width=0.127\textwidth,clip, trim=0cm 0cm 0cm 0.3cm]{images_2023/vgl_infer/mischa_second_capture/ours25/out_765_eye_[0.26344025, -0.16371429, 0].png} & 
		\includegraphics[frame,width=0.127\textwidth,clip, trim=0cm 0cm 0cm 0.3cm]{images_2023/vgl_infer/chris/ours25/out_297_eye_[0.3, -0.5, 0].png} & 
		\includegraphics[frame,width=0.127\textwidth,clip, trim=0cm 0cm 0cm 0.3cm]{images_2023/vgl_infer/chris/out_ours25_255_eye_[0.29093665, -0.16120547, 0].png} &
		\includegraphics[frame,width=0.127\textwidth,clip, trim=0cm 0cm 0cm 0.3cm]{images_2023/vgl_infer/chris/ours25/out_173_eye_[-0.5, 0.5, 0].png} &
		\includegraphics[frame,width=0.127\textwidth,clip, trim=0cm 0cm 0cm 0.3cm]{images_2023/vgl_infer/judge2/ours25/out_303_eye_[0.28455943, -0.20911498, 0].png} & 
		\includegraphics[frame,width=0.127\textwidth,clip, trim=0cm 0cm 0cm 0.3cm]{images_2023/vgl_infer/judge2/ours25/out_199_eye_[0.23014766, -0.2326444, 0].png} & \\
		|(out1_b)|\includegraphics[frame,width=0.127\textwidth,clip, trim=0cm 0cm 0cm 0.3cm]{images_2023/vgl_infer/mischa_second_capture/base+TCF/out_48_eye_[0.26344025, -0.17296118, 0].png} &
		\includegraphics[frame,width=0.127\textwidth,clip, trim=0cm 0cm 0cm 0.3cm]{images_2023/vgl_infer/mischa_second_capture/base+TCF/out_765_eye_[0.26344025, -0.16371429, 0].png} &
		\includegraphics[frame,width=0.127\textwidth,clip, trim=0cm 0cm 0cm 0.3cm]{images_2023/vgl_infer/chris/base+TCF/out_297_eye_[0.3, -0.5, 0].png} &
		\includegraphics[frame,width=0.127\textwidth,clip, trim=0cm 0cm 0cm 0.3cm]{images_2023/vgl_infer/chris/out_base+TCF_255_eye_[0.29093665, -0.16120547, 0].png} &
		\includegraphics[frame,width=0.127\textwidth,clip, trim=0cm 0cm 0cm 0.3cm]{images_2023/vgl_infer/chris/base+TCF/out_173_eye_[-0.5, 0.5, 0].png} &
		\includegraphics[frame,width=0.127\textwidth,clip, trim=0cm 0cm 0cm 0.3cm]{images_2023/vgl_infer/judge2/base+TCF/out_303_eye_[0.28455943, -0.20911498, 0].png} &
		\includegraphics[frame,width=0.127\textwidth,clip, trim=0cm 0cm 0cm 0.3cm]{images_2023/vgl_infer/judge2/base+TCF/out_199_eye_[0.23014766, -0.2326444, 0].png} &  \\[.0cm]
		|(out2)|\includegraphics[frame,width=0.127\textwidth,clip, trim=0cm 0cm 0cm 0.3cm]{images_2023/vgl_infer/mischa_second_capture/ours+5fix/out_48_eye_[0.26344025, -0.17296118, 0].png} &
		\includegraphics[frame,width=0.127\textwidth,clip, trim=0cm 0cm 0cm 0.3cm]{images_2023/vgl_infer/mischa_second_capture/ours+5fix/out_765_eye_[0.26344025, -0.16371429, 0].png} &
		\includegraphics[frame,width=0.127\textwidth,clip, trim=0cm 0cm 0cm 0.3cm]{images_2023/vgl_infer/chris/ours+5fix/out_297_eye_[0.3, -0.5, 0].png} &
		\includegraphics[frame,width=0.127\textwidth,clip, trim=0cm 0cm 0cm 0.3cm]{images_2023/vgl_infer/chris/out_ours+5fix_255_eye_[0.29093665, -0.16120547, 0].png} &
		\includegraphics[frame,width=0.127\textwidth,clip, trim=0cm 0cm 0cm 0.3cm]{images_2023/vgl_infer/chris/ours+5fix/out_173_eye_[-0.5, 0.5, 0].png} &
		\includegraphics[frame,width=0.127\textwidth,clip, trim=0cm 0cm 0cm 0.3cm]{images_2023/vgl_infer/judge2/ours+5fix/out_303_eye_[0.28455943, -0.20911498, 0].png} &
		\includegraphics[frame,width=0.127\textwidth,clip, trim=0cm 0cm 0cm 0.3cm]{images_2023/vgl_infer/judge2/ours+5fix/out_199_eye_[0.23014766, -0.2326444, 0].png} & \\
      };

      \begin{pgfonlayer}{background}
      \node[fill=blue!25,draw,rounded corners, inner sep= 1.5pt,fit=(in1)(m-4-2)] (b0) {};
      \node[fill=green!25,draw,rounded corners, inner sep= 1.5pt,fit=(in3)(m-4-5)(m-4-3)] (b1) {};
      \node[fill=blue!25,draw,rounded corners, inner sep= 1.5pt,fit=(in6)(m-4-7)(m-4-6)] (b2) {};
      \end{pgfonlayer}

      \node[rotate=90,anchor=south,inner sep=1pt] at ($(in1.west)+(-0.09,0)$) {\textbf{Input}\\VR / Eye};
      \node[rotate=90,anchor=south,inner sep=1pt] at ($(out1.west)+(-0.09,0)$) {\textbf{Ours}};
      \node[rotate=90,anchor=south,inner sep=1pt] at ($(out1_b.west)+(-0.09,0)$) {\textbf{Base~\cite{baseline}}};
      \node[rotate=90,anchor=south,inner sep=1pt] at ($(out2.west)+(-0.09,0)$) {\textbf{Ours}\\(5-Fix) };
    } \vspace{-4.5ex}
	\caption{ 
	Generated faces during inference, given mouth camera image and eye coordinates. See also the \href{https://www.ais.uni-bonn.de/videos/IROS_2023_Rochow/}{\textit{supplementary video}} for an animated comparison.}\vspace{-2ex}
	\label{fig:qualitative} \vspace{-1.5ex}
\end{figure*}

\subsection{Qualitative Results}
Qualitative results are shown in \cref{fig:q_gt,fig:q_fancy,fig:qualitative}. \cref{fig:q_gt} compares our method with ground truth and the baseline.
It shows exemplary results of our quantitative evaluation in \cref{tab:eval:ablations}.
As can be seen, our results are much more accurate and closer to the ground truth.
Unlike our method, the baseline fails whenever a bad expression image is retrieved.

\cref{fig:q_fancy} demonstrates very challenging mouth expressions obtained when mapping from the mouth camera input to a different person. This experiment shows that, unlike our baseline~\cite{baseline}, the proposed mouth camera guidance allows to resolve keypoint ambiguities and properly displays very challenging facial expressions, such as lips which partly stick together.

\cref{fig:qualitative} contains inference results compared with the baseline. 
In particular, we want to highlight that Ours-5-Fix produces almost the same results as our method configuration with image retrieval (Ours).
The supplementary video (see \cref{supplementary_video}) contains an animated comparison.

	\subsection{Throughput and Latency}
	We use pipelining techniques to enhance the throughput from 29\,fps to 34\,fps on an NVIDIA A6000 GPU with very low latency
	(34\,ms excluding and 51\,ms including camera exposure time).

    \subsection{The ANA Avatar XPRIZE Finals}
    At the ANA Avatar XPRIZE competition finals in November 2022, our team and three different operators had to accomplish three test runs, of which the first one was a qualification run.
    The goal was to complete ten different tasks as fast as possible. For each completed task one point was awarded.
    Five additional points were awarded for usability and the ability to understand emotions and gestures.
    Especially for these, a facial animation was mandatory.
    Two tasks consisted of interacting with a human recipient.
    Our facial animation pipeline allowed seamless and immersive interaction between operator and recipient, which was rewarded with a full judge score on all three days.
    Overall, our Team NimbRo achieved a perfect score (15/15) with the fastest time in all three runs.

	\section{Conclusion}
	We proposed a real-time capable VR facial animation approach that generalizes well to unseen operators and allows for modeling a broader range of facial expressions, compared to keypoint-driven approaches.
	We extended the baseline with a source image attention mechanism and developed a way to inject visual mouth image information into the animation pipeline without overfitting.
	These two extensions yield better accuracy and significantly improve temporal consistency which is important for smooth interaction.
	Our method still struggles in generating unusual expressions such as sticking out the tongue.
	Furthermore, movement in the upper face is still limited.
	\printbibliography

\end{document}